\documentclass[runningheads]{llncs}

\usepackage{epsfig}
\usepackage{graphicx}
\usepackage{booktabs} 
\usepackage{subfigure}
\usepackage{algorithm}
\usepackage{algorithmic}
\usepackage{xspace}
\usepackage{cite}

\usepackage{makecell}
\usepackage{xcolor}
\usepackage{multirow} 
\usepackage{bigstrut} 

\usepackage{tikz}
\usepackage{comment}
\usepackage{amsmath,amssymb} 
\usepackage{color}

\usepackage{pdfpages} 

\usepackage[accsupp]{axessibility}  

\usepackage[pagebackref=true,breaklinks=true,letterpaper=true,colorlinks,bookmarks=false]{hyperref}
\usepackage{bm} 

\def\bgamma{\mbox{{\boldmath $\gamma$}}}

\def\mL{{\mathcal L}}

\def\mN{{\mathcal N}}

\def\mP{{\mathcal P}}
\def\mQ{{\mathcal Q}}

\def\mS{{\mathcal S}}

\def\mV{{\mathcal V}}

\DeclareMathAlphabet\mathbfcal{OMS}{cmsy}{b}{n}

\def\0{{\bf 0}}
\def\1{{\bf 1}}

\def\bB{{\bf B}}

\def\bD{{\bf D}}

\def\bI{{\bf I}}

\def\bM{{\bf M}}

\def\bP{{\bf P}}

\def\bb{{\bf b}}

\def\bs{{\bf s}}
\def\bt{{\bf t}}

\def\bv{{\bf v}}

\def\mmR{{\mathbb R}}

\def\mmI{{\mathbb I}}

\def\bgamma{{\bm \gamma}}

\def\bb{{\bf b}}

\def\bP{{\bf P}}

\def\eg{\emph{e.g.,}} 
\def\ie{\emph{i.e.,}} 
 
\def\etc{\emph{etc.}} 
 
\def\etal{{\em et al.\/}\,}

\newcommand{\topline}{\toprule[0.1em]}
\newcommand{\midline}{\midrule[0.075em]}

\newcommand{\bottomline}{\bottomrule[0.1em]}

\usepackage{xspace}

\begin{document}
\pagestyle{headings}
\mainmatter
\def\ECCVSubNumber{3844}  

\setlength{\textfloatsep}{6pt}

\newcommand{\name}{Densely-Anchored Sampling\xspace}
\newcommand{\shortname}{DAS\xspace}
\newcommand{\lowername}{densely-anchored sampling\xspace}

\newcommand{\bfs}{Discriminative Feature Scaling\xspace}
\newcommand{\shortbfs}{DFS\xspace}

\newcommand{\tmb}{Memorized Transformation Shifting\xspace}
\newcommand{\shorttmb}{MTS\xspace}

\newcommand{\fullstop}{~}
\newcommand{\na}{\text{N/A}}


\newcommand{\crrevision}{\textcolor{black}}

\newcommand{\lec}{\textcolor{black}}
\newcommand{\vtars}{\textcolor{black}}
\newcommand{\xxl}{\textcolor{black}}
\newcommand{\xhm}{\textcolor{black}}
\newcommand{\zsh}{\textcolor{black}}
\newcommand{\ice}{\textcolor{black}}
\newcommand{\kui}{\textcolor{black}}
\newcommand{\revision}{\textcolor{black}}
\def\hd{\textcolor{black}}

\newcommand{\improve}{\textbf}
\newcommand{\important}{\textcolor{red}}

\definecolor{corrected}{RGB}{177,222,140}
\definecolor{wrong}{RGB}{255,51,0}

\title{\shortname: \name for Deep Metric Learning} 

\titlerunning{\shortname: \name for Deep Metric Learning}
%
\author{Lizhao Liu\inst{1,2} \and
Shangxin Hunag\inst{1} \and
Zhuangwei Zhuang\inst{1} \and \\
Ran Yang\inst{1} \and
Mingkui Tan\inst{1,3}$^\dag$ \and
Yaowei Wang\inst{2}\thanks{Corresponding authors.}}
\authorrunning{L. Liu et al.}
%
\institute{
$^1$South China University of Technology $^2$PengCheng Laboratory 
\\$^3$Information Technology R\&D Innovation Center of Peking University
\email{\{selizhaoliu,sevtars,z.zhuangwei,msyangran\}@mail.scut.edu.cn}\texttt{,}\\
\email{mingkuitan@scut.edu.cn}\texttt{,} \email{wangyw@pcl.ac.cn}
}



\makeatletter
\renewcommand*{\@fnsymbol}[1]{\ensuremath{\ifcase#1\or \dagger\or \ddagger\or
		\mathsection\or \mathparagraph\or \|\or **\or \dagger\dagger
		\or \ddagger\ddagger \else\@ctrerr\fi}}
\makeatother

\maketitle

\begin{abstract}
\kui{Deep Metric Learning (DML) serves to learn an embedding function to project semantically similar data into nearby embedding space and plays a vital role in many applications, such as image retrieval and face recognition. However, the performance of DML methods often highly depends on sampling methods to choose effective data from the embedding space in the training. In practice, the embeddings in the embedding space are obtained by some deep models, \lec{where the embedding space is often with} \ice{barren area} due to the absence of training points, resulting in so called ``missing embedding'' issue.}
\lec{This issue may impair the sample quality, which leads to degenerated DML performance. 
In this work, we investigate how to alleviate the ``missing embedding'' issue to improve the sampling quality and achieve effective DML. To this end, we propose a \name (\shortname) scheme that \revision{considers the embedding with corresponding data point as ``anchor'' and exploits the anchor's} nearby embedding space to densely produce embeddings without data points. Specifically, we propose to exploit the embedding space around single anchor with \bfs (\shortbfs) and multiple anchors with \tmb (\shorttmb). In this way, by combing the embeddings with and without data points, we are able to provide more embeddings to facilitate the sampling process thus boosting the performance of DML.
Our method is effortlessly integrated into existing DML frameworks and improves them without bells and whistles. Extensive experiments on three benchmark datasets demonstrate the superiority of our method.}
\end{abstract}

\begin{figure}[t]
    \centering
    \includegraphics[width=0.5\linewidth]{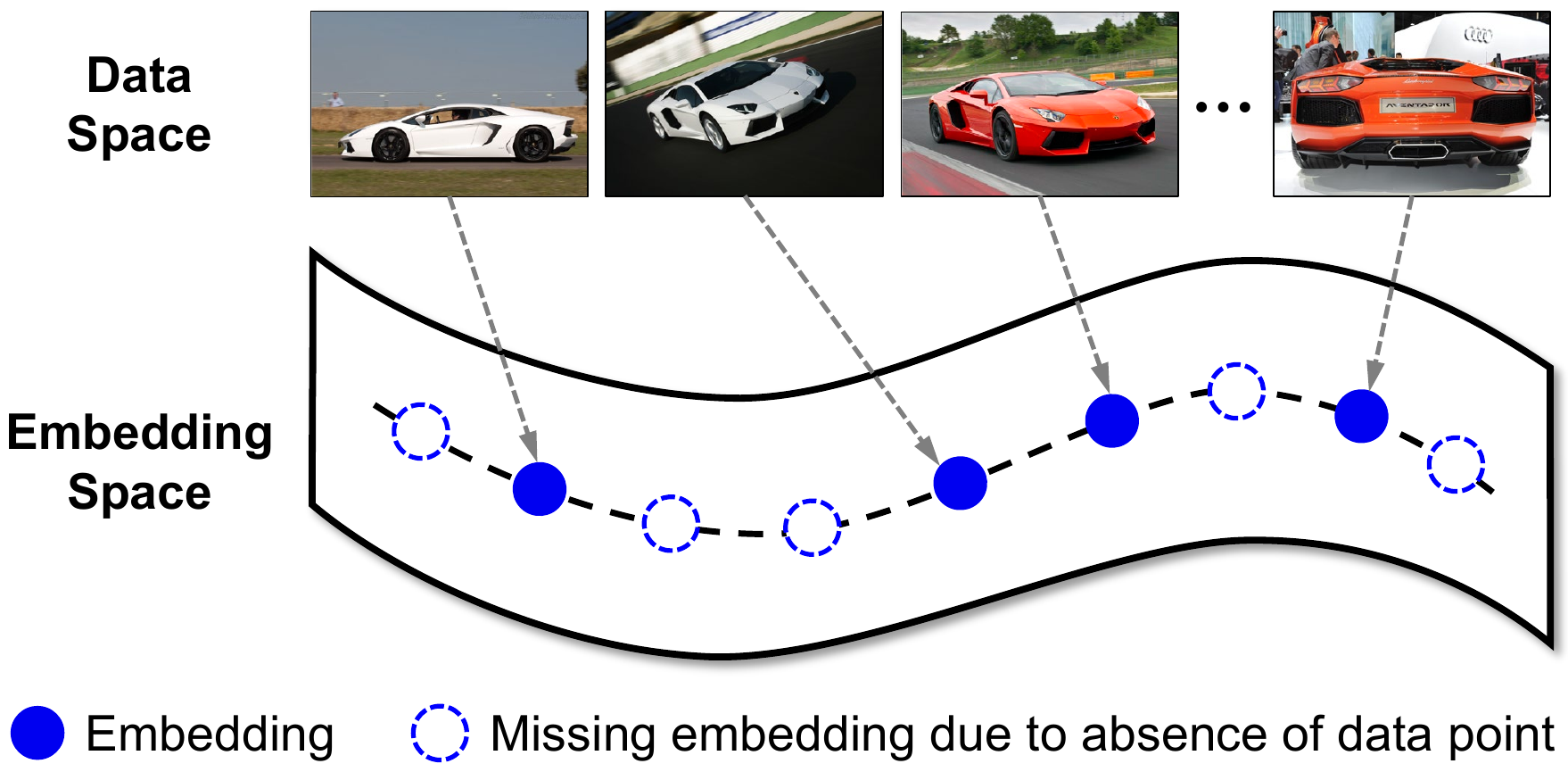}
    \caption{\lec{Illustration of the ``missing embedding'' issue. The data points of similar semantics are mapped into the \revision{nearby} embedding space \lec{that \revision{is} often with} \ice{barren area} due to the absence of data points,} resulting in the ``missing embedding'' issue\fullstop}
    \label{fig:motivation1}
\end{figure}

\section{Introduction}
Deep Metric learning (DML) is the foundation of various applications, including face recognition, verification~\cite{deng2019arcface,schroff2015facenet}, image retrieval~\cite{hoi2010semi}, image clustering~\cite{guo2017improved}, image classification~\cite{ding2016robust}, few-shot learning~\cite{liu2020dynamic}, \crrevision{video representation learning~\cite{chen2020rspnet} and sound generation~\cite{chen2020generating}} \etc~Since it was introduced, it has sparked considerable interest in the community, where academics have offered a variety of methods~\cite{hadsell2006dimensionality,movshovitz2017no,oh2016deep,schroff2015facenet,sohn2016improved,wang2019multi,wu2017sampling} and have made substantial progress~\cite{musgrave2020metric,roth2020revisiting}. The goal of DML is to learn \lec{a deep model} that is capable of mapping semantically similar \lec{data points} to \lec{similar embeddings in the embedding space.} To accomplish this, most existing approaches~\cite{chen2017beyond,hadsell2006dimensionality,schroff2015facenet,ustinova2016learning,wang2019multi,wu2017sampling} \lec{train the deep model with loss functions that bring the embeddings from semantically similar data points close to each other and vice versa. However, some embeddings may have limited contribution or bring no improvement to train the deep model~\cite{wu2017sampling}, or even lead to bad local minima early on in training (such as a collapsed model)~\cite{schroff2015facenet}. Thus, sampling informative and stable embeddings is very important to facilitate the training of deep model~\cite{wu2017sampling}.} As a result, improving the sample quality is of significance \lec{to achieve effective DML}. There are two commonly used measures for this goal: designing more effective sampling methods or providing more \lec{embeddings}.

Pioneering efforts have made substantial progress toward the design of effective sampling methods upon embedding pairs~\cite{roth2020revisiting,schroff2015facenet,wu2017sampling} or a full batch of embeddings~\cite{movshovitz2017no,oh2016deep}. \lec{These methods typically perform sampling on a batch of embeddings, which often leads to inaccurate sampling results due to the following reasons.}
First, the batch size is typically constrained by the memory of a single GPU as the sampling process typically cannot cross different GPU devices~\cite{roth2020revisiting}. Second, even with GPU that has sufficient memory to support a larger batch size, \lec{the embedding space that contains the embeddings embedded by deep models may still with \ice{barren area} due to the absence of data points, resulting in a ``missing embedding'' issue} (as shown in Fig.~\ref{fig:motivation1}). \lec{Thus, the limited amount of embeddings may impair the sample quality and the performance of DML.} \lec{Based on the above analyses, we ask: ``Can we overcome the inaccurate sampling issue brought by the absence of data points?''}

Very recently, a few attempts~\cite{duan2018deep,gu2021proxy,ko2020embedding,wang2020cross,zheng2019hardness} have been committed to answering this question by pseudo embedding generation. Hard \lec{example} generation \lec{approaches}~\cite{duan2018deep,zheng2019hardness} generate hard embeddings from easy embeddings with an additional generative adversarial network or auto-encoder. Embedding expansion~\cite{ko2020embedding} performs interpolation between embeddings to achieve augmentation in embedding space. Cross batch memory~\cite{wang2020cross} maintains embeddings from previous iterations and considers them are still informative in the current batch in terms of sampling. However, these approaches either leverage additional sub-network~\cite{duan2018deep,zheng2019hardness}, which introduce extra training cost, or need further modification to the sampling and loss computation process~\cite{ko2020embedding,wang2020cross} in standard DML~\cite{hadsell2006dimensionality,hermans2017defense,schroff2015facenet,sohn2016improved,wang2019multi}, which may limit their applicability to other tasks.

\lec{In this paper, we seek to densely produce embeddings without data points to alleviate the ``missing embedding'' issue. In this way, with the combination of embeddings with and without data points, we are able to provide more embeddings for sampling to improve the sample quality and achieve effective DML.} Our motivation stems from a fundamental hypothesis of metric learning: \textit{the embeddings that are close to each other in the embedding space have similar semantics.} \revision{Unfortunately, how to exploit the embedding space to produce effective embeddings without data points remains an unsolved problem.} \lec{\revision{To this end}, we propose a \name (\shortname) scheme to \revision{consider the embedding with data point as ``anchor'' and densely} exploit \revision{the anchor's} nearby embedding space to produce embeddings that have no corresponding data points. \revision{The proposed \shortname is consist of} two modules, namely, \bfs (\shortbfs) and \tmb (\shorttmb), which exploit the embedding space around single and multiple embeddings respectively to produce effective embeddings with no corresponding data points. To be specific, based on observations that effective semantics for one embedding are highly activated features~\cite{bau2017network,bau2020understanding,escorcia2015relationship}, \shortbfs identifies these features and applies random scaling on them. In this sense, we are able to exploit the embedding space around a single embedding by enhancing or weakening its effective semantics to produce embeddings. 
Based on the fact that semantic differences (\ie~transformations) of intra-class embeddings can be added to other embeddings to generate effective embeddings~\cite{lin2018deep}, we assume that they can be added in a way like word embeddings~\cite{mikolov2013linguistic}: \crrevision{\textit{$\text{Queen} = \text{Woman} + (\text{King} - \text{Man})$}}. Thus, our \tmb module exploits the embedding space among multiple embeddings by adding (\ie~shift) intra-class embeddings' semantic differences to other embeddings of the same class to produce effective embeddings. }

Our main contributions are summarized as follows. {\textbf{First}}, \lec{we propose a novel and \crrevision{plug-and-play} \name (\shortname) scheme \crrevision{that exploits embeddings' nearby embedding space and densely produces embeddings without data points to improve the sampling quality and performance of DML.}} \textbf{Second}, \lec{we propose \crrevision{two modules, namely} \bfs (\shortbfs) \crrevision{and \tmb (\shorttmb)} to exploit embedding space around a single \crrevision{and multiple} embeddings to produce embeddings.} \crrevision{\textbf{Last}, extensive experiments demonstrate the effectiveness of the proposed method.} 
\section{Related Work}
\noindent\textbf{Sampling Methods in DML.} \lec{Sampling informative and stable embeddings is vital to train the deep model in DML~\cite{roth2020revisiting,schroff2015facenet,wu2017sampling}.} Thus, various sampling approaches~\cite{ge2018deep,roth2020revisiting,schroff2015facenet,wang2019multi,wu2017sampling} have been tailored \lec{to effectively sample the embeddings to train the deep model}.
To further improve sampling efficiency, some researchers propose to leverage a whole batch of embeddings~\cite{hermans2017defense,kim2020proxy,movshovitz2017no,wang2019multi,zhu2020fewer}. Even though sophisticated sampling methods improve DML, \lec{due to the absence of data points, sampling embeddings that are often with ``missing embedding'' leads to inaccurate sampling, thereby degenerating the final performance.} \lec{In this paper, we propose a \shortname scheme to produce \lec{embeddings with no data points by exploiting embeddings' nearby embedding space} to achieve effective DML.}

\noindent\textbf{Loss Functions for DML.} \lec{Studies on DML losses} can be grouped into two categories: pair-based and proxy-based. The pair-based losses~\cite{hadsell2006dimensionality,hu2014discriminative,kim2019deep,schroff2015facenet,sohn2016improved,wang2014learning,wang2019ranked,wang2019multi,wu2017sampling,yu2019deep} are constructed upon the pairwise distance between embeddings. Although pair-based approaches mine the rich information among vast embedding pairs, they typically encounter the sampling effective embedding pairs issues. Instead, proxy-based~\cite{aziere2019ensemble,deng2019arcface,kim2020proxy,liu2017sphereface,movshovitz2017no,qian2019softtriple,zhai2018classification} losses introduce the concept of ``proxy'' as a class representation and avoid the sampling issue by optimizing the embedding close to its proxy. \revision{However, proxy-based methods are very difficult to train when the number of classes is extremely large~\cite{oh2016deep}, limiting their applicability to real-life scenarios. Thus, in this paper, we focus on developing an effective technique to alleviates the general data sampling issue for pair-based methods that have wider applicability and delivers boosted performance for them.}


\noindent\textbf{Pseudo Embedding Generation}. \lec{Methods on synthesizing pseudo embedding have been recently shown as an important technique to improve DML~\cite{duan2018deep,ko2020embedding,lin2018deep,wang2020cross,zhao2018adversarial,zheng2019hardness}.}
\crrevision{DAML~\cite{duan2018deep} uses an additional generative adversarial network to generates only hard negatives to improve the model training.
HDML~\cite{zhao2018adversarial} leverages the {inter-class} information for embedding generation on the sampled embeddings.
DVML~\cite{lin2018deep} apply extra generator and decoder to model the class centers that may have inaccurate distance estimation to the real embeddings and employs them to generate embeddings.}
Embedding expansion~\cite{ko2020embedding} linearly interpolates the embeddings to obtain more embeddings. Cross batch memory~\cite{wang2020cross} stores embeddings from previous batches and considers them beneficial for the upcoming sampling process. However these approaches~\cite{ko2020embedding,wang2020cross} suffer from \crrevision{the following} limitations: \textbf{First}, the generated embeddings can only be considered as negative during sampling, which may \lec{limit the power of them}. \textbf{Second}, the sampling or loss computation process need to be modified to dock with them, \crrevision{limiting their applicability.} \crrevision{\textbf{Third}, an additional sub-network is introduced in order to generate embeddings, which brings heavy computation cost. \textbf{Last}, information source that may have inaccurate distance measurement to real embeddings such as class centers and inter-class differences are considered to produce embeddings.} \crrevision{Different from} them, \shortname \crrevision{is a light-weight module, which produces embeddings by densely sampling around the ``anchor'' and serves as a plug-and-play component to facilitate the sampling process in standard DML.}

\noindent\textbf{Data Augmentation.} \lec{Augmentation in data space such as image~\cite{shorten2019survey} has been widely studied and is considered an important technique to avoid overfitting.} Recently, many efforts have been made to design effective augmentation methods~\cite{devries2017dataset,chu2020feature,volpi2018adversarial,wang2019implicit,yin2019feature} in feature space, aiming to provide more features for training when the source data (\eg~images) is scarce. These methods often introduce complicated training processes~\cite{chu2020feature,volpi2018adversarial}. \revision{\crrevision{Wang~\etal~\cite{wang2019implicit} propose ISDA that} estimates semantic differences with covariance matrices and develops an improved version of cross-entropy loss. The computation and memory consumption of covariance matrices are \crrevision{much} heavier than \shortname. Moreover, \crrevision{ISDA can not be trivially extended to pair-based DML loss.} \crrevision{Yin~\etal~\cite{yin2019feature} introduce FTL that} requires extra networks (\eg~decoder and feature transfer module) and a carefully designed bi-stage training strategy to achieve feature translation, which is less efficient and general than \shortname.} Unlike these methods, we view the feature space augmentation as an approach to fill the ``missing \revision{embedding}'' in the \revision{embedding} space and propose a simpler solution under the context of DML.

\begin{figure*}[!t]
    \centering
    \includegraphics[width=1.0\linewidth]{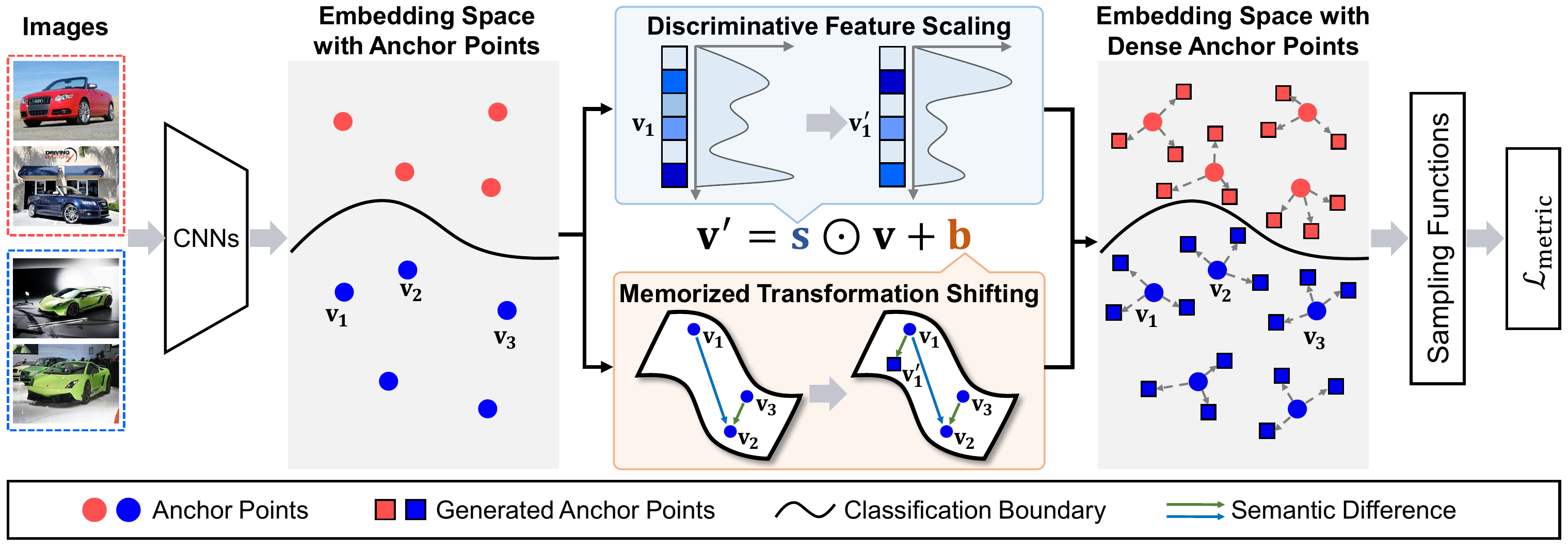}
    \caption{\lec{Illustration of our \name (\shortname) scheme, which leverages two modules to exploit \revision{anchors' (\ie~embeddings with data points)} nearby embedding space to densely produce embeddings without data points:} \shortbfs performs random scaling on the discriminative features to \lec{produce embeddings} around a single embedding; \shorttmb \lec{exploits the embedding space among multiple embeddings by} adding the intra-class semantic differences to embeddings of the same class to \lec{produce embeddings.} With \shortname, we \lec{alleviate the ``missing embedding'' issue by providing more embeddings for sampling, thereby, achieving effective DML\fullstop} 
    }
    \label{fig:overall}
\end{figure*}

\section{\name}

\crrevision{\lec{In this paper, we seek to improve DML by alleviating the ``missing embedding'' issue incurred in sampling.} The overall process of \shortname scheme is shown in Fig.~\ref{fig:overall}.}

\noindent\textbf{Notation}. Let $\bv = f_{\theta}(\bI) \in \mmR^d$ be an embedding \lec{with data point $\bI$, where} $f_{\theta}(\cdot)$ is a deep model (\eg~CNNs) with learnable parameters $\theta$. Let $y_{\bv}$ be the label of the embedding $\bv$. \lec{Let $\bv'$ denote embedding without data point.} Following previous methods~\cite{schroff2015facenet}, we normalize both $\bv$ and $\bv'$ to the $d$-dimensional hyper-sphere (\ie~$\| \bv \|_2, \| \bv' \|_2 = 1$). We omit the normalization process for brevity. Let $C$ denote the number of training classes.

\subsection{Problem Definition and Motivation}
\label{sec:overall}
\lec{DML seeks to learn a deep model that keeps similar data points close, and vice versa. Formally, we define the distance between two embeddings as follows~\cite{wu2017sampling}:
\begin{equation}
    \bD_{ij} = \| f_{\theta}(\bI_i) - f_{\theta}(\bI_j) \|,
\end{equation}
where $\|\cdot\|$ denotes the $\ell_2$ norm. For any positive pair of embeddings ($y_{i} = y_{j}$), the distance should be small; Whilst for negative pair ($y_i \neq y_j$), it should be large. In practice, limited by computing resources, it is infeasible to optimize every element in $\bD_{ij}$. Therefore, it is necessary to sample effective embedding pair for the objective construction (take the contrastive loss as an example):
\begin{equation}
    \mL = \sum_{(i,j) \in \mQ} \mmI\{y_i = y_j\}~\bD_{ij} + \mmI\{y_i \neq y_j\}~[\alpha - \bD_{ij}]_{+},
\end{equation}
where $\mmI\{\cdot\}$ is the indication function, $\alpha$ is a margin, $\mQ = \operatorname{S}(D)$ indicates indexes of the sampled embedding \revision{pairs} and $\operatorname{S}$ denotes some sampling function.
}

\ice{Without causing ambiguity, we denote the points on the embedding space as ``\textbf{anchor points}'', based on which we will conduct sampling \lec{to train the deep model}. Note that each \lec{data} point shall have an anchor point on the embedding space. However, due to the absence of training data, the embedding space may have a lot of ``barren area''.}
\lec{
Thus, it is very difficult to provide sufficient anchor points for sampling and learn a deep model with good performance. To this end, we propose to densely produce anchor points with no data points to facilitate the sampling, \revision{thereby improving the training} in DML. 
}

Specifically, we \revision{first} propose to exploit the embedding space around a single embedding by enhancing or weakening its semantics. We term this process as semantic scaling.
\revision{Second}, we propose to add (\ie~shift) intra-class differences (\ie~transformations) to embeddings to exploit the embedding space among multiple embeddings. We term this process as semantic shifting. Based on the above analyses, the formulation of \shortname scheme is formulated as
\begin{equation}
    \label{eqn:esc}
    \bv' = \operatorname{\shortname}(\bv;~\bs, \bb) = \underbrace{\bs~\odot~\bv}_\emph{scaling}\underbrace{+~~\bb}_\emph{shifting},
\end{equation}
where $\bv$ and $\bv'$ are embeddings with and without data points, respectively. $\odot$ denotes the Hadamard product. $\bs, \bb \in \mmR^d$ are semantic scaling and semantic shifting factors, respectively. Moreover, given a set of semantic scaling and shifting factor pairs $\{(\bs_t, \bb_t)\}$, we are able to produce a set of embeddings by
\begin{equation}
    \label{eqn:esc_multi}
    \bv'_t = \operatorname{\shortname}(\bv;~\bs_t, \bb_t).
\end{equation}

\crrevision{Therefore,} the semantic scaling and shifting factors is essential to the quality of the \lec{produced} embeddings and we propose \shortbfs and \shorttmb to \crrevision{acquire} them:
\begin{align}
    \bs &= \operatorname{\shortbfs}(\{\bv \mid y_{\bv} = c \}), \\
    \bb &= \operatorname{\shorttmb}(\{\bv \mid y_{\bv} = c \}),
\end{align}
where \shortbfs and \shorttmb take embeddings from the same class as input and produce the semantic scaling and shifting factors, respectively.

\subsection{\bfs}
\label{sec:bfs}

\begin{figure*}[!h]
    \centering
    \begin{minipage}[t]{.48\linewidth}
        \centering
        \includegraphics[width=\textwidth]{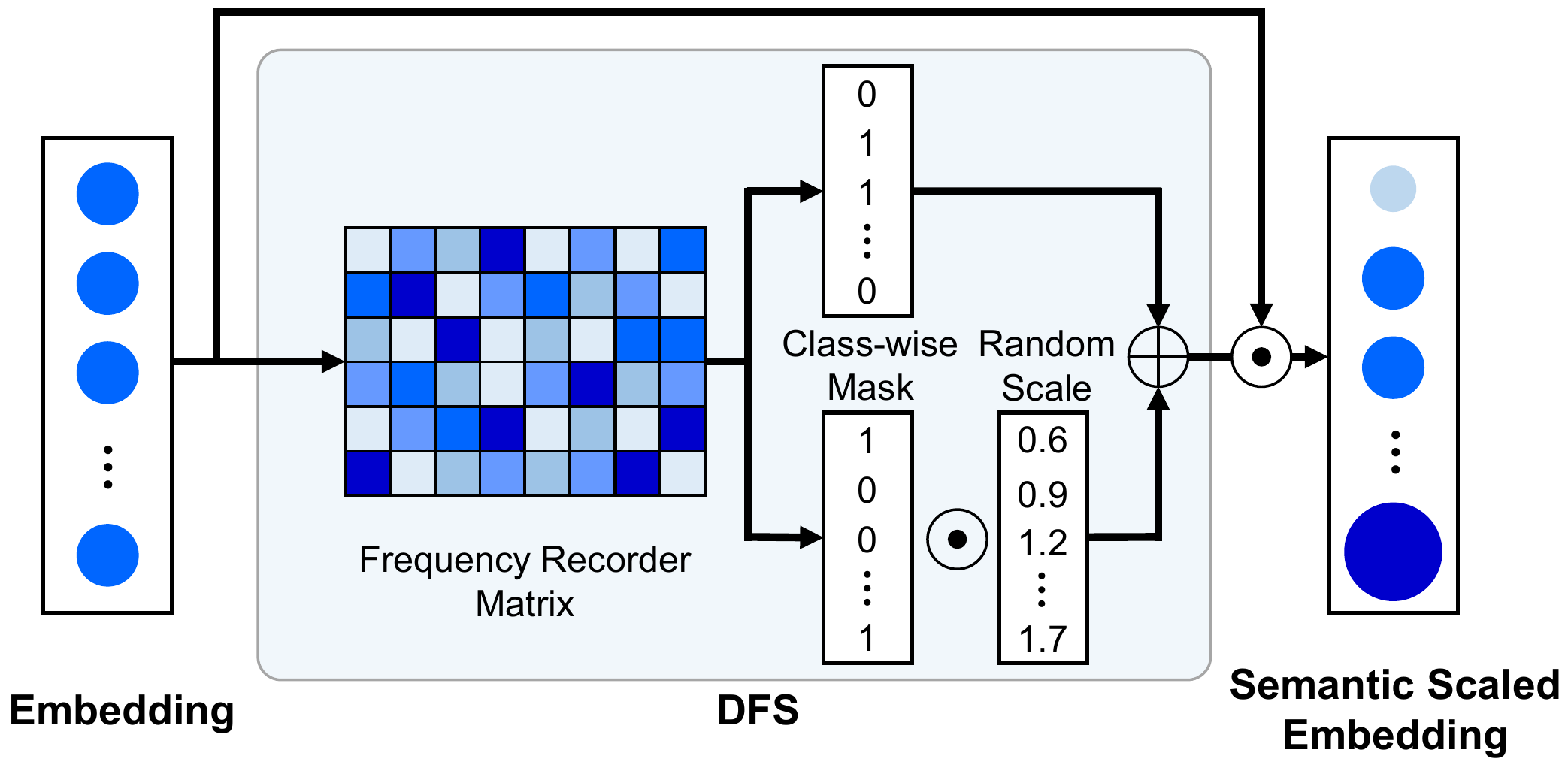}
        \caption{Illustration of the proposed \bfs (\shortbfs) module, which identifies the discriminative features (\eg~channels) and applies different random scaling to them to produce embeddings around a single embedding\fullstop 
        }
        \label{fig:bfs}
    \end{minipage}
    \hskip 0.08in
    \begin{minipage}[t]{.48\linewidth}
        \centering
        \includegraphics[width=\textwidth]{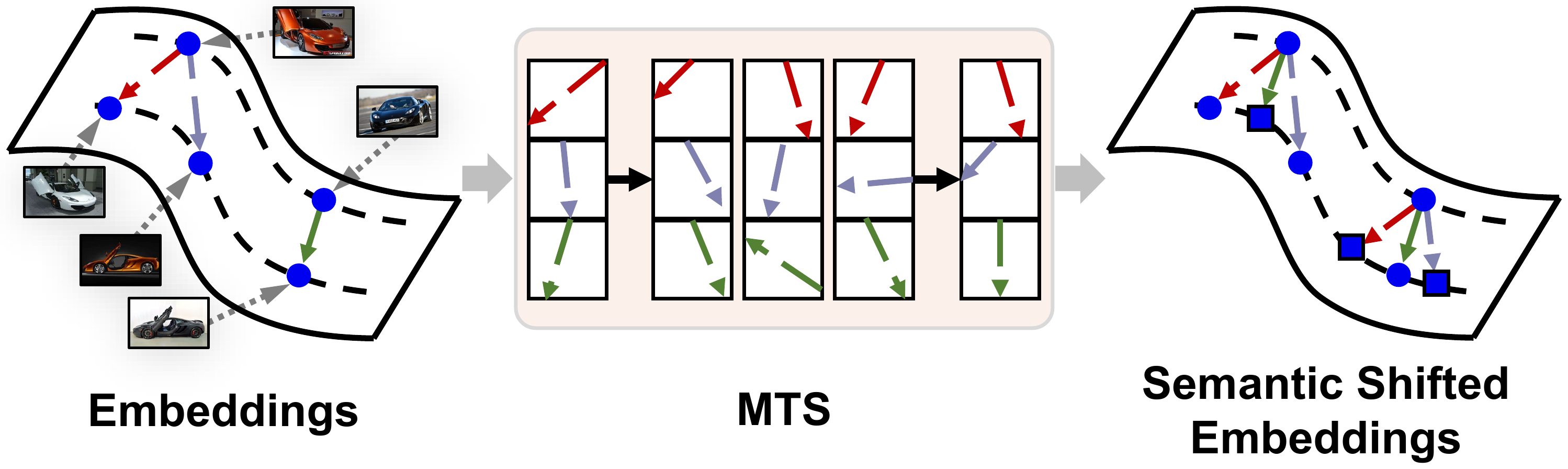}
        \caption{Illustration of the proposed \tmb (\shorttmb) module. \shorttmb adds the intra-class transformations to embeddings of the same class to \lec{produce embeddings among multiple embeddings\fullstop}}
        \label{fig:tmb}
    \end{minipage}
\end{figure*}


We seek to obtain effective semantic scaling factors to \lec{produce anchor points} around a single anchor. Thus, as shown in Fig.~\ref{fig:bfs}, our \bfs (\shortbfs) module carries out the semantic scaling mechanism by finding the discriminative features in an embedding and applying random scaling on them. The feasibility of \shortbfs comes from two aspects. First, visual attributes can be predicted reliably using a sparse number of neurons from CNNs~\cite{escorcia2015relationship}. Second, in CNNs, neurons that match a diverse set of object concepts are highly activated~\cite{bau2017network,bau2020understanding}. In practice, the embedding typically occupies a high dimensional embedding space, which contains semantics for all training classes and semantics for each class are diverse. Thus, the semantics for one class are more likely to be noise for another. In this sense, it is very important to find out the effective \eg~discriminative features for each class in order to perform semantic scaling. To this end, we propose to identify the effective semantics by counting the number of occurrences of the highly-activated neurons for each class.

To be specific, we first initialize a Frequency Recorder Matrix (FRM) $\bP \in \mmR^{C \times d}$ as a zero matrix. Then, given a set of embeddings $\{\bv \mid y_{\bv} = c\}$ from class $c$, we update $\bP$ as follows:
\begin{equation}
    \label{eqn:frm}
    \bP[c, k] = 
    \left\{
    \begin{aligned}
        & \bP[c, k] + 1, &\text { if } \bv[k] \in \operatorname{Top}(\bv, K), \\
        & \bP[c, k], &\text { otherwise, }
    \end{aligned}
    \right.
\end{equation}
where $\operatorname{Top}(\bv, K)$ is an operator to select the top-$K$ elements from the vector $\bv$. During training, we constantly update the FRM by recording the position of highly-activated neurons from the embeddings of the same class. In this way, FRM\footnote{Visualization of the frequency recorder matrix is in the supplementary.} serves as a stable, accurate \crrevision{and} effective semantics identifier for \crrevision{training} classes. Given the FRM, we compute the
class-wise binary channel mask $\bM \in \mmR^{C \times d}$ by
\begin{equation}
    \label{eqn:mask}
    \bM[c, k] = 
    \left\{
    \begin{aligned}
        & 1, & \text { if } \bP[c, k] \in \operatorname{Top}(\bP[c], K), \\
        & 0, & \text { otherwise. }
    \end{aligned}
    \right.
\end{equation}
Given an embedding $\bv$, we attain the semantic scaling factor by
\begin{equation}
    \label{eqn:scaling_factors}
    \bs =  \bgamma \odot \bM[y_{\bv}] \revision{~+~\mathbf{1}^d \odot (1 - \bM[y_{\bv}])},
\end{equation}
where $\bgamma \in \mmR^d$ and $\bgamma \sim \operatorname{Uniform}[1 - r_s, 1 + r_s]^d$ and $r_s \in (0, 1)$ is a hyper-parameter to be set. \revision{Note that we only randomly scale the discriminative features while leaving the indiscriminative ones intact.} To produce more than one scaling factor, we \revision{repeatedly} sample $\bgamma$ from the uniform distribution. 

\subsection{\tmb}
\label{sec:tmb}


\lec{To produce anchors without data points among multiple anchors, we propose a module to provide effective semantic shifting factors. As shown in Fig.~\ref{fig:tmb}, our \tmb (\shorttmb) module}
exploit intra-class embeddings' nearby embedding space by leveraging the differences between embeddings and adding them to other embeddings of the same class. On the basis of that semantic differences of embeddings \lec{can be added to other embedding to generate effective embeddings}~\cite{lin2018deep}, the motivation of our \shorttmb comes from the semantic relations of word embedding~\cite{mikolov2013linguistic}: \crrevision{\textit{$\text{Woman} + (\text{King}-\text{Man}) =\text{Queen}$}}, where a ``woman'' pluses ``royal'' semantics (\eg~transformation) becomes a ``Queen''.

The transformations from both inter-class and intra-class embeddings are candidates for our design choices. However, 
the majority of the inter-class transformations typically are not transferable due to large inter-class differences. Thus, we only consider intra-class embeddings to attain the transformations. As suggested by the latest research~\cite{roth2020revisiting}, sampling only two images for each class in a batch consistently achieves good performance, \lec{which} leads to very limited transformations we can obtain in one batch \ie~two transformations. To address this issue, we \lec{construct} a bank to memorize the transformations from previous iterations to ensure the diversity of the transformations. Specifically, we construct a transformation bank $\bB \in \mmR^{C \times Z \times d}$, where $Z$ is the bank capacity for each class. Then, during training, once we obtain a set of embeddings $\mV_c = \{\bv \mid y_{\bv} = c\}$ from the class $c$, we calculate the transformations between them by
\begin{equation}
    \label{eqn:transformation}
    \bt_{z} = \bv_i - \bv_j,~~\bv_i, \bv_j \in \mV_c, i \neq j.
\end{equation}
Then, the transformations are \crrevision{en-queued} into $\bB$ according to the \crrevision{FIFO} principle to ensure that the transformations in the bank are in a relatively fresh state:
\begin{equation}
    \label{eqn:fifo}
    \bB[y_{\bv}, z] = \bt_{z}, ~z \in \{1, 2, \dots, Z\}.
\end{equation}
Note that $z$ is reset to 1 when it reaches $Z$. Finally, with the assistance of the bank $\bB$, given an embedding v, we retrieve the semantic shifting factor as follow
\begin{equation}
    \label{eqn:shifting}
    \bb = r_b \bt, ~~ \bt \sim \{\bB[y_{\bv}, z]~\mid~z \small{=} {1, 2, \dots, Z}\}.
\end{equation}
$r_b$ is a hyper-parameter. Multiple shifting factors \crrevision{are formed by repeat sampling.}

\begin{algorithm}[t]
    \footnotesize{
    \caption{Training method of \shortname-based DML}
    \label{alg:esc_based_dml}
    \begin{algorithmic}[1]
    \REQUIRE
    Training image-label pairs $\mS = \{(\bI_i, y_i)\}_{i=1}^{N}$;
    the embedding function $f_{\theta}$;
    number of embeddings to produce $T$;
    number of training classes $C$;
    transformation bank capacity $Z$; 
    learning rate $\alpha$. \\
    \ENSURE
    Optimized embedding function $f_{\theta}^{*}$. \\
    \STATE Initialize $\theta$ from ImageNet pretrained model. 
    
    \STATE Initialize the frequency recorder matrix $\bP \in \mmR^{C \times d} = \textbf{0}$. \\
    \STATE Initialize the transformation bank $\bB \in \mmR^{C \times Z \times d} = \textbf{0}$. \\
    
    \WHILE{not converge}
        \STATE Obtain a batch image-label pairs $\{(\bI_i, y_i)\}_{i=1}^{B}$ from $\mS$.
        \STATE Compute embeddings $\bv_i \leftarrow f_{\theta}(\bI_i), i = 1, 2, \cdots, B$.
        
        \STATE // \textbf{\emph{{perform semantic scaling \crrevision{by \bfs}}}}
        \STATE Update the frequency recorder matrix $\bP$ by Eqn.~(\ref{eqn:frm}). \\
        \STATE Acquire semantic scaling factors $\{\bs_j\}_{j=1}^{B \times T}$ by Eqn.~(\ref{eqn:scaling_factors}). \\
        \STATE // \textbf{\emph{{perform semantic shifting \crrevision{by \tmb}}}}
        \STATE Obtain intra-class transformations $\{\bt\}$ by Eqn.~(\ref{eqn:transformation}). \\
        \STATE Update the transformation bank $\bB$ by Eqn.~(\ref{eqn:fifo}). \\
        \STATE Attain semantic shifting factors $\{\bb_j\}_{j=1}^{B \times T}$ by Eqn.~(\ref{eqn:shifting}). \\
        \STATE // \textbf{\emph{{perform \lowername}}}
        \STATE Produce embeddings $\{\bv_{j}' \}_{j=1}^{B \times T}$ by Eqn.~(\ref{eqn:esc_multi}). \\
        
        \STATE Sample positive and negative embedding sets. \\

        \STATE Compute the training loss $\mL_\text{\crrevision{\shortname-DML}}$ by Eqn.~(\ref{eqn:loss}). \\
        \STATE Update the parameters $\theta$ by $ \theta \leftarrow \theta - \alpha\nabla_{\theta} \mL_\text{\crrevision{\shortname-DML}}$. \\
    \ENDWHILE
    \end{algorithmic}
    }
\end{algorithm}

\subsection{DML with \name}
The overall algorithm of \lec{integrating \shortname into DML} is detailed in Algorithm~\ref{alg:esc_based_dml}. Given \crrevision{anchor}-label pairs $\{(\bv, y_{\bv})\}$, we produce embedding-label pairs $\{(\bv', y'_{\bv})\}$ with no data points by \shortname scheme, where \crrevision{$y'_{\bv} = y_{\bv}$ since the class semantic is preserved.}
Then, embedding-label pairs with or without data points are fed into the sampling module to obtain the positive and negative embedding sets (\eg~pairs, triplets, \etc~\crrevision{specified by} the \crrevision{sampling} and loss functions): $\{(\mP, \mN)\} = \operatorname{Sample}(\{(\bv, y_{\bv})\} \cup \{(\bv', y'_{\bv})\})$. \crrevision{Last}, given a DML loss function $\mL_\text{DML}$\footnote{
See supplementary for detailed DML sampling methods and loss functions.
}, \shortname-based DML objective function is formulated as:
\begin{equation}
    \label{eqn:loss}
    \mL_\text{\crrevision{\shortname-DML}} = \mL_\text{DML}(\{(\mP, \mN)\}).
\end{equation}



\begin{table*}[!ht]\small
    {
    \caption{Comparisons with \crrevision{SoTA} methods \crrevision{on} CUB, CARS and SOP. The best results are in \textbf{bold}. $*$~indicates the reimplementation by~\cite{ko2021learning}. \dag~denotes our reimplementation\fullstop
	\label{tab:sota}
	}
    \resizebox{\linewidth}{!}{
	\begin{tabular}{lc|cccc|cccc|cccc}
		\toprule
		\multirow{2}{*}{Method} & \multirow{2}{*}{Backbone} & \multicolumn{4}{c|}{CUB} & \multicolumn{4}{c|}{CARS} & \multicolumn{4}{c}{SOP} \\ \cline{3-14} 
		& & R@1 & R@2 & R@4 & R@8 & R@1 & R@2 & R@4 & R@8 & R@1 & R@10 & R@100 & R@1000  \\
        \midrule
		Margin~\cite{wu2017sampling} & $\operatorname{R}^{128}$ & 63.60 & 74.40 & 83.10 & 90.00 & 79.60 & 86.50 & 91.90 & 95.10 & 72.70 & 86.20 & 93.80 & 98.00 \\
        HDC~\cite{yuan2017hard} & $\operatorname{G}^{384}$ & 53.60 & 65.70 & 77.00 & 85.60 & 73.70 & 83.20 & 89.50 & 93.80 & 69.50 & 84.40 & 92.80 & 97.70 \\
        A-BIER~\cite{opitz2018deep} & $\operatorname{G}^{384}$ & 57.50 & 68.70 & 78.30 & 86.20 & 82.00 & 89.00 & 93.20 & 96.10 & 74.20 & 86.90 & 94.00 & 97.80 \\
        ABE~\cite{kim2018attention} & $\operatorname{G}^{512}$ & 60.60 & 71.50 & 79.80 & 87.40 & 85.20 & 90.50 & 94.00 & 96.10 & 76.30 & 88.40 & 94.80 & 98.20 \\
        HTL~\cite{ge2018deep} & $\operatorname{IBN}^{512}$ & 57.10 & 68.80 & 78.70 & 86.50 & 81.40 & 88.00 & 92.70 & 95.70 & 74.80 & 88.30 & 94.80 & 98.40 \\
        RLL-H~\cite{wang2019ranked} & $\operatorname{IBN}^{512}$ & 57.40 & 69.70 & 79.20 & 86.90 & 74.00 & 83.60 & 90.10 & 94.10 & 76.10 & 89.10 & 95.40 & \na \\
        SoftTriple~\cite{qian2019softtriple} & $\operatorname{IBN}^{512}$ & 65.40 & 76.40 & 84.50 & 90.40 & 84.50 & 90.70 & 94.50 & 96.90 & 78.30 & 90.30 & 95.90 & \na \\
        MS~\cite{wang2019multi} & $\operatorname{IBN}^{512}$ & 65.70 & 77.00 & 86.30 & 91.20 & 84.10 & 90.40 & 94.00 & 96.50 & 78.20 & 90.50 & 96.00 & 98.70 \\
        ProxyGML~\cite{zhu2020fewer} & $\operatorname{IBN}^{512}$ & 66.60 & 77.60 & 86.40 & \na & 85.50 & 91.80 & 95.30 & \na & 78.00 & 90.60 & 96.20 & \na \\
        ProxyAnchor~\cite{kim2020proxy} & $\operatorname{IBN}^{512}$ & 68.40 & 79.20 & 86.80 & 91.60 & 86.10 & 91.70 & 95.00 & 97.30 & 79.10 & 90.80 & 96.20 & 98.70 \\
        \midrule
        Contrastive + XBM~\cite{wang2020cross} & $\operatorname{IBN}^{512}$ & 65.80 & 75.90 & 84.00 & 89.90 & 82.00 & 88.70 & 93.10 & 96.10 & 79.50 & 90.80 & 96.10 & 98.70 \\
        MS$^*$~\cite{wang2019multi} & $\operatorname{IBN}^{512}$ & 64.50 & 76.20 & 84.60 & 90.50 & 82.10 & 88.80 & 93.20 & 96.10 & 76.30 & 89.70 & 96.00 & 98.80 \\
        MS + EE$^*$~\cite{ko2020embedding} & $\operatorname{IBN}^{512}$ & 65.10 & 76.80 & 86.10 & 91.00 & 82.70 & 89.20 & 93.80 & 96.40 & 77.00 & 89.50 & 96.00 & 98.80 \\
        ProxyAnchor + MemVir~\cite{ko2021learning} & $\operatorname{IBN}^{512}$ & 69.00 & 79.20 & 86.80 & 91.60 & 86.70 & 92.00 & 95.20 & 97.40 & 79.70 & 91.00 & 96.30 & 98.60 \\
        \midrule
        MS$^\dag$~\cite{wang2019multi} & $\operatorname{IBN}^{512}$ & 65.72 & 77.19 & 85.74 & 91.56 & 83.86 & 90.41 & 94.64 & 96.99 & 76.89 & 89.58 & 95.59 & 98.60 \\
        MS + \shortname($K=8$) (Ours) & $\operatorname{IBN}^{512}$ & 67.07 & 78.11 & 86.43 & 91.88 & 85.66 & 91.60 & 95.27 & 97.37 & 78.16 & 90.26 & 95.99 & 98.76 \\
        MS$^\dag$~\cite{wang2019multi} & $\operatorname{R}^{512}$ & 66.46 & 77.28 & 85.85 & 91.69 & 83.99 & 90.39 & 94.51 & 96.80 & 79.53 & 91.06 & 96.30 & 98.83 \\
        MS + \shortname($K=8$) (Ours) & $\operatorname{R}^{512}$ & \textbf{69.19} & \textbf{79.25} & \textbf{87.09} & \textbf{92.62} & \textbf{87.84} & \textbf{93.15} & \textbf{95.99} & \textbf{97.85} & \textbf{80.59} & \textbf{91.80} & \textbf{96.68} & \textbf{98.95} \\
        \bottomrule
	\end{tabular}
	}
	}
	\centering
	\vskip 0.08in
\end{table*}

\section{Experiments}

\noindent\textbf{Datasets.} We \crrevision{use} three \crrevision{popular} benchmarks: 1) {CUB2011-200 (CUB)}~\cite{wah2011caltech}, a fine-grained bird dataset with the first 100 categories for training and another 100 categories for testing. 2) {CARS196 (CARS)}~\cite{krause20133d}, a fine-grained vehicle dataset \crrevision{with} the first 98 classes \crrevision{for training} and another 98 classes \crrevision{for testing}. 3) {Stanford Online Products (SOP)}~\cite{oh2016deep}, a large-scale online products dataset \crrevision{with} the train and test partitions as 11,318 classes and another 11,316 classes, respectively.

\noindent\textbf{Implementation details\footnote{See supplementary for more details.}} We leverage two popular backbones: ResNet50~\cite{he2016deep} ($\operatorname{R}^{d}$) and Inception BN~\cite{ioffe2015batch} ($\operatorname{IBN}^{d}$), where their parameters are initialized from ImageNet~\cite{deng2009imagenet} pre-trained models and $d$ denotes the embedding dimension. Note that some approaches also consider GoogleNet~\cite{szegedy2015going} ($\operatorname{G}^{d}$) as the backbone. Here, we mainly consider the settings of $d = 128, 512$. $\operatorname{R}^{128}$ is used as the default backbone. The embedding layer is randomly initialized. Regarding evaluation metrics, Recall at k (R@k)~\cite{jegou2010product}, Normalized Mutual Information (NMI)~\cite{schutze2008introduction} and F1 score (F1)~\cite{sohn2016improved} are used, where R@k measures the image retrieval performance while F1 and NMI measure the image clustering performance. For hyper-parameters in \shortname, we set $(T, K, Z, r_s, r_b) = (3, 4, 10, 1\text{e}^{-2}, 1\text{e}^{-2})$ by default.\footnote{Experiments on hyper-parameters $T, r_s, r_b$ are in the supplementary.} \crrevision{Our source code is publicly available at \url{https://github.com/lizhaoliu-Lec/DAS}.}

\begin{table*}[ht!]
    \centering
    \caption{Comparisons with pair-based methods on CUB, CARS and SOP. \crrevision{[S] and [D] denote semi-hard and distance weighted sampling, respectively\fullstop}}
	\label{tab:improvement}
    \resizebox{0.75\linewidth}{!}{
	\begin{tabular}{l|ccc|ccc|ccc}
		\toprule
		\multirow{2}[0]{*}{Method} & \multicolumn{3}{c|}{CUB} & \multicolumn{3}{c|}{CARS} & \multicolumn{3}{c}{SOP} \\ \cline{2-10} 
		& R@1 & F1 & NMI & R@1 & F1 & NMI & R@1 & F1 & NMI \\
		\midrule
		Triplet [S]~\cite{schroff2015facenet} & 60.25 & 32.82 & 64.64 & 74.64 & 31.98 & 63.22 & 73.51 & 33.47 & 89.33 \\
		Triplet [S] {+ \shortname} & \textbf{60.82} & \textbf{33.86} & \textbf{65.67} & \textbf{77.21} & \textbf{33.88} & \textbf{64.84} & \textbf{73.99} & \textbf{33.91} & \textbf{89.42}  \\
		\midrule
        Triplet [D]~\cite{wu2017sampling} & 62.68 & 36.39 & 67.03 & 78.86 & 35.80 & 65.85 & 77.54 & 37.10 & 90.05 \\
		Triplet [D] {+ \shortname} & \textbf{64.28} & \textbf{38.16} & \textbf{68.06} & \textbf{82.63} & \textbf{39.14} & \textbf{68.12} & \textbf{77.95} & \textbf{37.64} & \textbf{90.18} \\
		\midrule
		Contrastive [D]~\cite{wu2017sampling} & 61.65 & 35.23 & 66.58 & 76.03 & 32.77 & 64.09 & 73.13 & 35.60 & 89.78 \\
		Contrastive [D] {+ \shortname} & \textbf{63.67} & \textbf{36.25} & \textbf{67.15} &
		\textbf{80.74} & \textbf{36.07} & \textbf{65.93} & \textbf{74.80} & \textbf{36.21} & \textbf{89.89} \\
		\midrule
		Margin~\cite{wu2017sampling} & 62.61 & 37.33 & 67.58 &
		80.10 & 37.85 & 67.15 & 78.69 & 39.20 & 90.50 \\
		Margin {+ \shortname} & \textbf{64.50} & \textbf{37.86} & \textbf{68.04} &
		\textbf{82.29} & \textbf{38.22} & \textbf{67.94} & \textbf{79.14} & \textbf{39.52} & \textbf{90.56} \\ 
		\midrule
		GenLifted~\cite{hermans2017defense} & 58.81 & 34.64 & 65.50 & 72.45 & 32.43 & 64.00 & 76.18 & 37.26 & 90.13 \\
		GenLifted {+ \shortname} & \textbf{59.94} & \textbf{35.09} & \textbf{66.07} &
		\textbf{73.55} & \textbf{32.85} & \textbf{64.11} & \textbf{76.92} & \textbf{37.64} & \textbf{90.21} \\
		\midrule
		N-Pair~\cite{sohn2016improved} & 60.55 & 36.94 & 67.19 & 77.35 & 36.26 & 66.74 & 77.71 & 37.13 & 90.15 \\
		N-Pair {+ \shortname} & \textbf{62.81} & \textbf{38.37} & \textbf{68.43} & \textbf{79.93} & \textbf{38.06} & \textbf{68.20} & \textbf{77.98} & \textbf{37.82} & \textbf{90.28} \\
		\midrule
		MS~\cite{wang2019multi} & 62.63 & 38.88 & 68.19 & 82.04 & 40.85 & 69.45 & 78.89 & 37.53 & 90.12 \\
		MS {+ \shortname} & \textbf{64.13} & \textbf{39.18} & \textbf{69.08} &
		\textbf{83.31} & \textbf{42.78} & \textbf{70.77} & \textbf{79.44} & \textbf{38.77} & \textbf{90.40} \\
    \bottomrule
	\end{tabular}
	}
\end{table*}

\subsection{Comparison with State-of-the-arts}
\label{sec:exp_sota}

In this section, we compare our method with state-of-the-art competitors to investigate the effectiveness of \shortname. The results are shown in Table~\ref{tab:sota}. For fair comparisons, the results of the closely related baseline EE are from the reimplementation by~\cite{ko2021learning} using the stronger $\operatorname{IBN}^{512}$ backbone ($\operatorname{G}^{512}$ in the original paper). Our approach is based on MS loss and achieves superior performance on all datasets and evaluation metrics. \textbf{First}, when combining the $\operatorname{R}^{512}$ backbone and \shortname, we are able to boost the R@1 metrics by $3.49\%$ on CUB and $1.74\%$ on CARS, which shows that \shortname is able to deliver more accurate image retrieval results even \lec{with higher embedding dimension (\ie~512)}. \textbf{Second}, for our closely relative opponent, EE, its improvements on MS are marginal, showing that simply performing interpolation to generate embeddings is inferior to \shortname. \textbf{Last}, even for a strong baseline, ProxyAnchor that leverages the advanced training techniques, and sophisticated loss, we still outperform it considerably.

\subsection{Effectiveness of \shortname on Pair-based Loss}
\label{sec:exp_pair}

\noindent\textbf{Quantitative results.}~To investigate the efficacy of \shortname, we conduct experiments \crrevision{with $\operatorname{R}^{128}$ backbone} on the widely used pair-based losses. We reimplement all baselines under the same settings for a fair comparison. The results are presented in Table~\ref{tab:improvement}. For considered pair-based losses, \shortname is able to improve their performance on both image retrieval and clustering metrics. Notably, for approaches such as GenLifted and N-Pair that leverage the whole batch of embeddings for loss computation, \shortname still improves their performance, showing its effectiveness. Last, even for a very strong baseline, MS, that considers different kinds of relationships among embedding pairs and designs sophisticated weighting mechanism, \shortname is able to greatly improve it without bells and whistles.

\begin{figure}
    \makebox[\linewidth][c]{%
    \begin{subfigure}
    \centering
    \includegraphics[width=.40\linewidth]{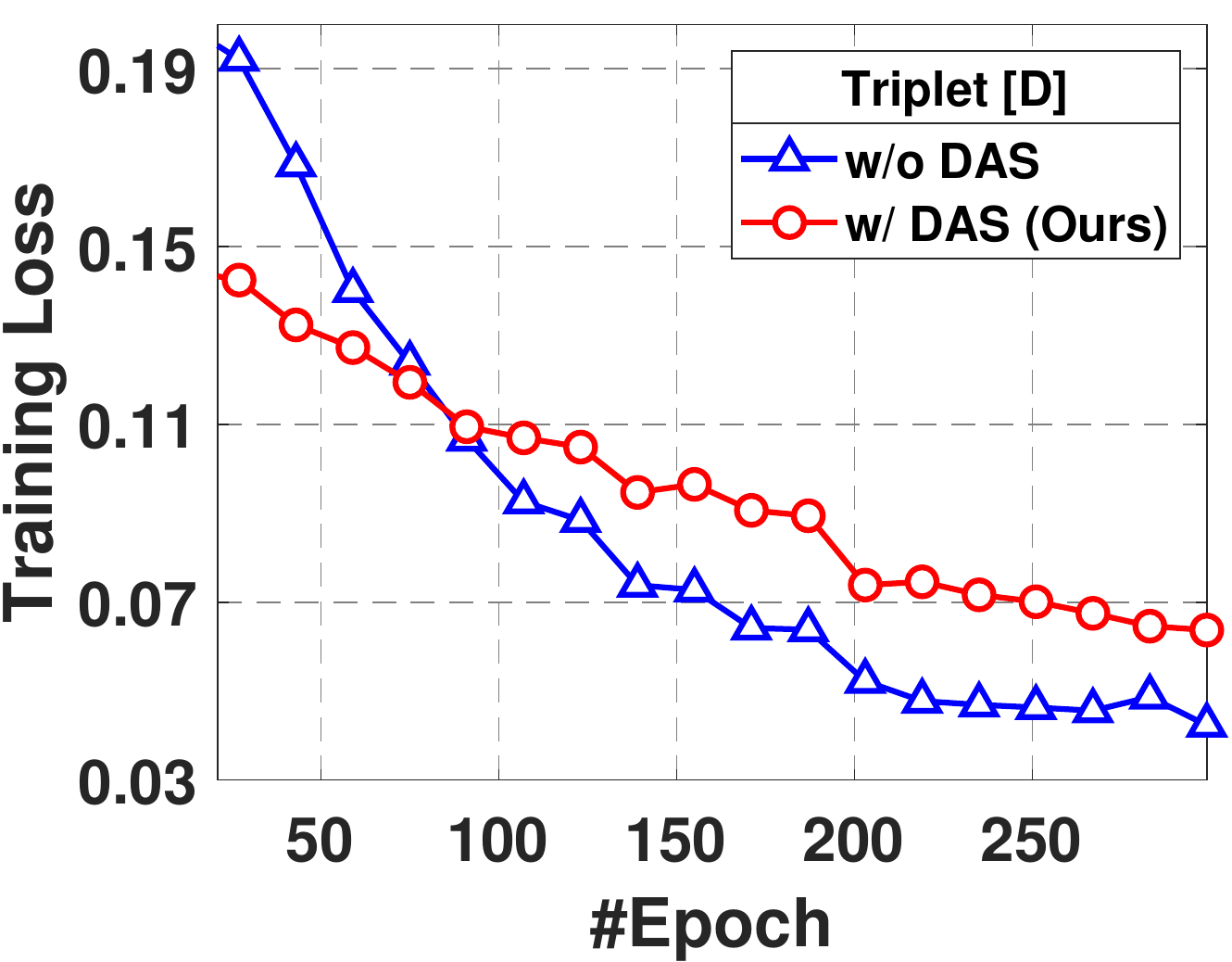}
    \end{subfigure}%
    
    \begin{subfigure}
    \centering
    \includegraphics[width=.40\linewidth]{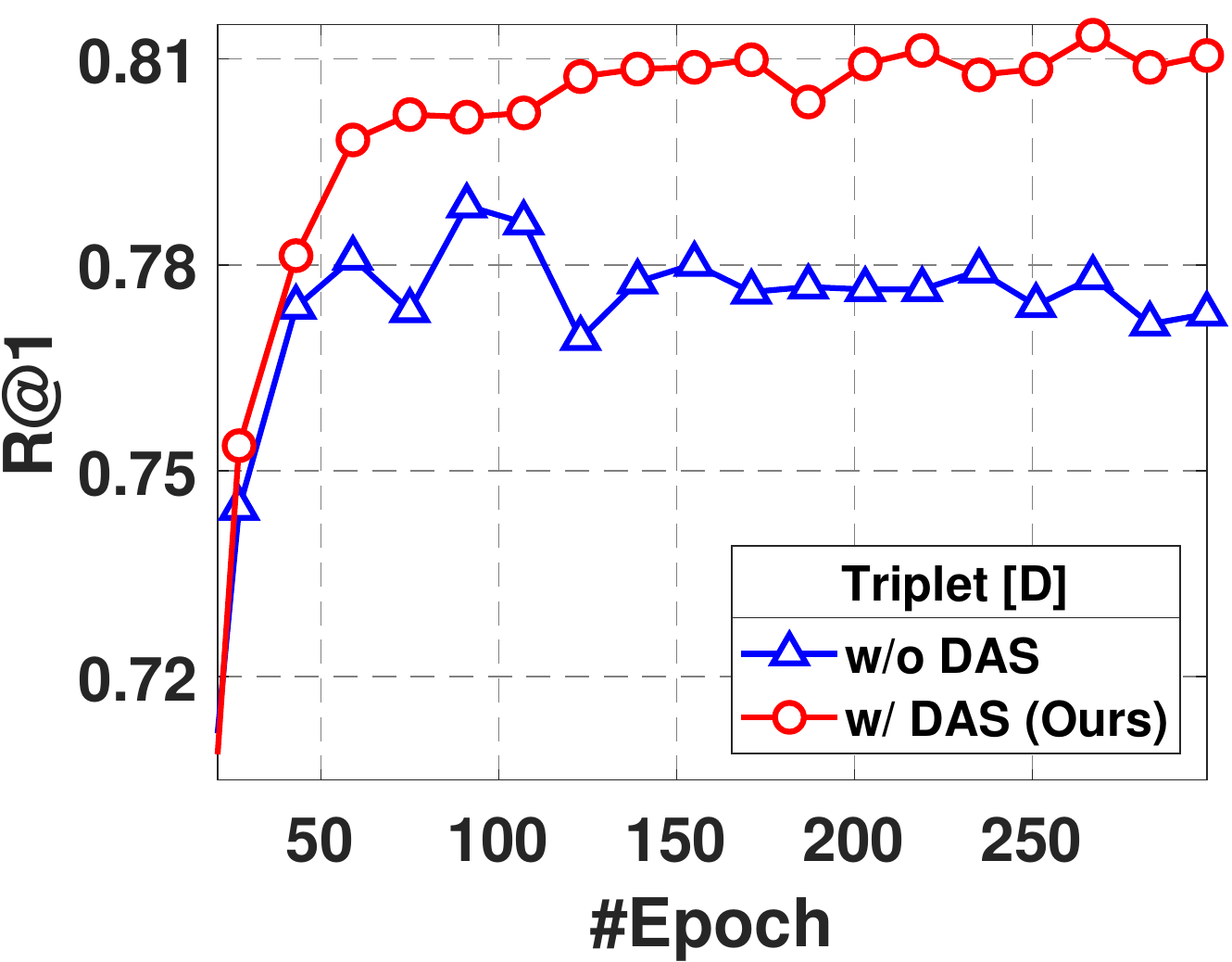}
    \end{subfigure}%
    }
    \caption{The training loss and test set R@1 on CARS. The sampling method and loss function are triplet loss and distance weighted sampling, respectively\fullstop}
    \label{fig:training}
\end{figure}

\noindent\textbf{Convergence analyses.}~We provide the results of training loss and test set R@1 in Fig.~\ref{fig:training} to analyze the training behaviors of \shortname.\footnote{Results on more pair-based losses are in the supplementary.} The loss curve with \shortname decreases smoother than that without \shortname, showing that \shortname provides more embeddings to facilitate sampling, thereby, stabilizing the training. Moreover, with \shortname, the training loss is higher than the baseline, one possible reason is that \shortname is able to act as a regularizer to avoid overfitting, which is consistent with the result that \shortname achieves a higher test set R@1. Similar phenomenon is also observed in other embeddings generation methods~\cite{ko2020embedding}.

\begin{table*}[!h]
    \centering
    \begin{minipage}[t]{.47\linewidth}
        \centering
        \caption{Comparisons with various sampling methods on CARS\fullstop}
        \label{tab:different_sampling}
        \resizebox{\textwidth}{!}{
        \begin{tabular}{llcccc}
            \toprule
    		Method & Sampling & \shortname & R@1 & F1 & NMI \\
            \midrule
    		\multirow{8}{*}{Triplet} & \multirow{2}{*}{Random} &  & 74.21 &33.41 & 64.28 \\
    		& & \checkmark & \textbf{76.79} & \textbf{35.21} & \textbf{65.49} \\ \cline{2-6} 
    		&  \multirow{2}{*}{Semi-hard~\cite{schroff2015facenet}} &  & 74.64 & 31.98 & 63.22 \\
    		& & \checkmark & \textbf{77.10} & \textbf{33.82} & \textbf{65.03} \\ \cline{2-6} 
    		& \multirow{2}{*}{Soft-hard~\cite{roth2020revisiting}} & & 79.20 & 35.55 & 66.08 \\
    		& & \checkmark & \textbf{80.54} & \textbf{37.52} & \textbf{66.77} \\ \cline{2-6} 
    		& \multirow{2}{*}{Distance~\cite{wu2017sampling}} & & 78.86 & 35.80 & 65.85 \\
    		& & \checkmark & \textbf{81.34} & \textbf{37.27} & \textbf{67.21} \\
    		\midrule
    		\multirow{4}{*}{Contrastive} & \multirow{2}{*}{Random} &  & 42.44 & 15.83 & 48.87 \\
    		& & \checkmark & \textbf{50.79} & \textbf{19.40} & \textbf{52.71} \\ \cline{2-6} 
    		& \multirow{2}{*}{Distance~\cite{wu2017sampling}} & & 76.03 & 32.77 & 64.09 \\
    		& & \checkmark & \textbf{80.70} & \textbf{35.47} & \textbf{66.01} \\
    		\bottomrule
	    \end{tabular}
	    }
	    
    \end{minipage}
    \hskip 0.02in
    \begin{minipage}[t]{.50\linewidth}
        \centering
        \caption{Comparisons with more related works on CARS\fullstop}
    	\label{tab:sota}
    	\vspace{0.03in}
        \resizebox{\linewidth}{!}{
        \begin{tabular}{l|c|cccc}
            \topline
    		Method & Backbone & R@1 & R@2 & R@4 & R@8 \\
            \midline
            N-Pair + HDML~\cite{zheng2021hardness} & $\operatorname{G}^{512}$ & 68.90 & 78.90 & 85.80 & 90.90 \\
            N-Pair + HDML-A~\cite{zheng2021hardness} & $\operatorname{G}^{512}$ & 81.10 & 88.80 & 93.70 & 96.70 \\
            N-Pair + \shortname (Ours) & $\operatorname{G}^{512}$ & \textbf{83.70} & \textbf{90.33} & \textbf{94.47} & \textbf{96.77} \\ 
            \midline
            MS + SEC~\cite{zhang2020deep} & $\operatorname{IBN}^{512}$ & 85.73 & 91.96 & 95.51 & 97.54 \\
            MS + \shortname (Ours) & $\operatorname{IBN}^{512}$ & 85.66 & 91.60 & 95.27 & 97.37 \\
            MS + SEC + \shortname (Ours) & $\operatorname{IBN}^{512}$ & \textbf{87.80} & \textbf{93.16} & \textbf{96.18} & \textbf{98.01} \\
            \midline
            Margin + DiVA~\cite{milbich2020diva} & $\operatorname{IBN}^{512}$ & 83.10 & 90.00 & \na & \na \\
            Margin + \shortname (Ours) & $\operatorname{IBN}^{512}$ & \textbf{84.85} & \textbf{90.32} & {93.99} & {96.40} \\
            \midline
            ProxyNCA++~\cite{teh2020proxynca++} & $\operatorname{R}^{512}$ & 86.50 & 92.50 & 95.70 & 97.70 \\
            Margin + DiVA & $\operatorname{R}^{512}$ & 82.20 & 89.00 & \na & \na \\
            Margin + DRML~\cite{zheng2021deepiccv} & $\operatorname{R}^{512}$ & 73.30 & 83.00 & 89.80 & 94.40 \\
            Margin + DCML~\cite{zheng2021deepcvpr} & $\operatorname{R}^{512}$ & 85.20 & 91.80 & \textbf{96.00} & \textbf{98.00} \\
            Margin + \shortname (Ours) & $\operatorname{R}^{512}$ & \textbf{88.34} & \textbf{93.21} & {95.92} & {97.59} \\
            \bottomline
    	\end{tabular}
    	}
    \end{minipage}
    \vskip 0.08in
\end{table*}

\subsection{Effectiveness of \shortname on Sampling Method}
\label{sec:exp_sample}
In this section, we investigate the effectiveness of \shortname by evaluating it with different sampling approaches. We choose two \crrevision{popular} loss functions: triplet and contrastive losses \crrevision{that} are sensitive to the sampling methods. \crrevision{Therefore, } various sampling approaches \crrevision{are} tailored for them. The experiment results are presented in Table~\ref{tab:different_sampling}. For two loss functions, despite the choice of sampling approaches, \shortname is able to improve them considerably on both image retrieval and clustering metrics. Notably, when applying \shortname, triplet loss with random sampling outperform the one with the semi-hard sampling. This indicates that producing more embeddings for sampling is as important as the sampling approach. 

\subsection{Qualitative Results}
\label{sec:exp_qualitative}
\begin{figure*}[!h]
    \centering
    \includegraphics[width=\linewidth]{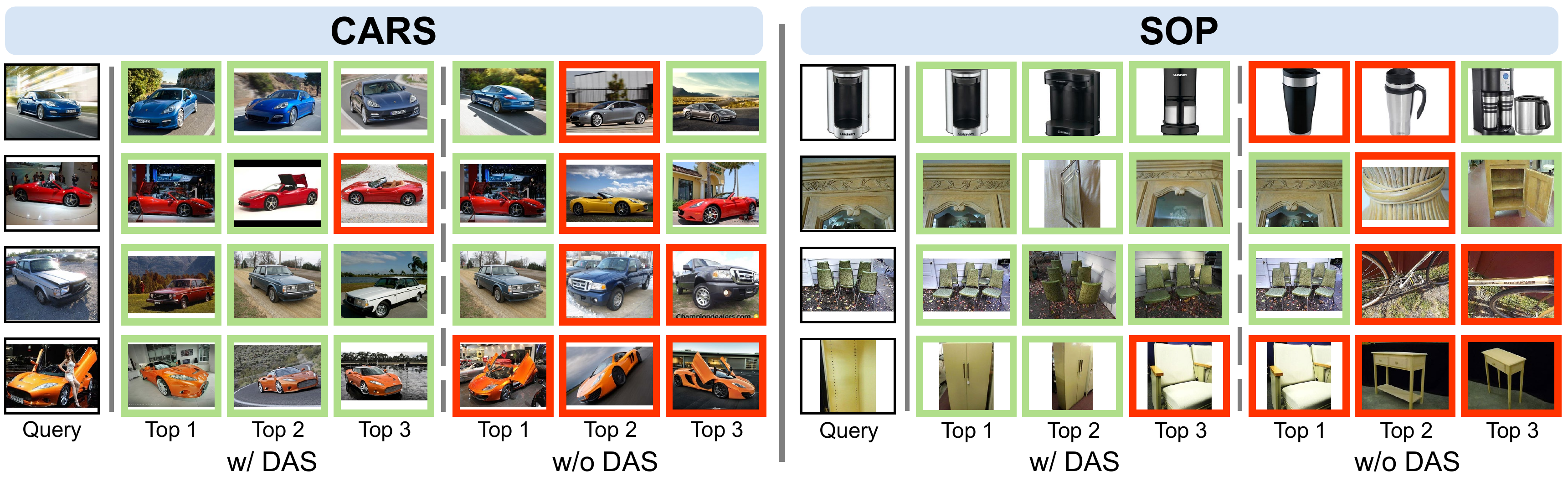}
    \caption{Top 3 retrieved results from the $\operatorname{R}^{128}$ trained w/ or w/o \shortname. \textcolor{corrected}{green} and \textcolor{wrong}{red} rectangles indicate desired and undesired results, respectively\fullstop}
    \label{fig:cars&online_compare}
\end{figure*}

To better understand our method, we compare contrastive loss (with distance weighted sampling) w/ or w/o \shortname and visualize image retrieval results on both CARS (Fig.~\ref{fig:cars&online_compare}) and SOP (Fig.~\ref{fig:cars&online_compare}) datasets.\footnote{More qualitative results are in the supplementary.} On CARS, the model performs better with \shortname despite the background noises or the interference from the car's color, demonstrating that \shortname enforces the model to focus on real semantics. On SOP, the model with \shortname is insensitive to drastic viewpoint changes. These results verify the generalization ability and robustness of \shortname under various scenes.

\subsection{Ablation Studies}
\label{sec:exp_ablation}
{\parindent0pt

\revision{\noindent{\textbf{Effect of \shortbfs and \shorttmb.}}} In this section, we perform ablation studies to evaluate the performance gain by each module in \shortname. The loss function and sampling method are triplet loss and distance weighted sampling, respectively. The results are in Table~\ref{tab:ablation}. First, \shortbfs, alone, boosts R@1 by +2.42\%, verifying that producing embeddings around a single embedding is able to force the model to focus on real semantics and achieve better image retrieval results. Second, with \shorttmb only, the clustering metrics are greatly improved, indicating that producing embeddings among multiple embeddings are beneficial to the image clustering task. Last, with \shortbfs and \shortname, all metrics are further improved, suggesting that \shortbfs and \shorttmb reinforce and complement each other.

\begin{table*}
    \centering
    \caption{Ablation studies on CARS\fullstop}
    \label{tab:ablation}
    \resizebox{0.55\textwidth}{!}{
        \begin{tabular}{cclll}
            \toprule
	        \shortbfs & \shorttmb & R@1 & F1 & NMI \\
            \midrule
	        &  &  78.86 & 35.80 & 65.85 \\
	        \checkmark &  &  81.28 \improve{(+2.42)} & 36.22 \improve{(+0.42)} & 66.84 \improve{(+0.99)} \\
	        & \checkmark &  81.83 \improve{(+2.97)} & 38.26 \improve{(+2.46)} & 67.81 \improve{(+1.96)} \\
	        \checkmark & \checkmark &  82.63 \improve{(+3.77)} & 39.14     \improve{(+3.34)} & 68.12 \improve{(+2.27)} \\
            \bottomrule
        \end{tabular}
    }
\end{table*}

\revision{\noindent{\textbf{Effect of $K$ in \shortbfs.}} Multi-dimensional embeddings have a diverse set of semantic features~\cite{bau2017network,bau2020understanding}. The top-K mask is effective to discover the discriminative features.
We perform experiments on SOP, a large-scale and diverse dataset with margin loss to support our claim. With {\small{$K=1,2,4,8,16,32$}}, we obtain results {\small{$\text{R@1} = 78.76, 78.86, \textbf{79.14}, 78.09, 77.94, 77.96$}}, where substantial improvements are observed  when $K=\{1,2,4\}$ and larger $K~(\text{\ie}~K > 4)$ leads to worse results.} 

\revision{\noindent{\textbf{Effect of $Z$ in \shorttmb.}} 
The larger bank capacity ($Z$) allows us to access intra-class transformations from current and previous iterations, which improves the diversity of the produced embeddings. We conduct experiments on CARS with margin loss by setting {\small{$Z=1,2,3,4,5$}} and obtain {\small{$\text{R@1}=82.35,82.38,82.42,\textbf{82.75},82.33$}}, showing that history transformations ({\small{$Z>2$}}) bring slight improvements.}
}

\subsection{Further Discussions}

\noindent{\crrevision{\textbf{More discussions on \shortbfs.} One may question whether the proposed \shortbfs requires the linear assumption on high dimensional feature space. In fact, we do not make this assumption and our method is built on a very basic hypothesis of metric learning: the embeddings that are close to each other in the embedding space have similar semantics. More critically, \shortbfs does not rely on the linearity assumption. Instead, \shortbfs is based on the observations that effective semantics for one embedding are highly activated features~\cite{bau2017network,bau2020understanding}.
}}

\noindent{\crrevision{\textbf{More discussions on \shorttmb.} Li~\etal~\cite{li2019memory} also apply a memory module is leveraged to store abundant features and conduct neighborhood search upon them to enhance the discriminative power of a general CNN feature on the image search and few-shot learning tasks. Unlike them, \shortname constructs a memory bank with the intra-class embedding transformations, which allows us to access intra-class transformations from current and previous iterations, and thus improves the diversity of produced embeddings for DML.}}

\noindent\textbf{Comparisons with more related works.} \vtars{In Table~\ref{tab:sota}, we compare \shortname with HDML~\cite{zheng2021hardness}, SEC~\cite{zhang2020deep}, DiVA~\cite{milbich2020diva}, ProxyNCA++~\cite{teh2020proxynca++}, DRML~\cite{zheng2021deepiccv}, DCML~\cite{zheng2021deepcvpr} on CARS under the same settings. We can see that \shortname outperforms HDML, DiVA, DRML and DCML, and achieves comparable performance to SEC. Note that \shortname can be also incorporated into SEC. As a result, using MS loss, SEC + \shortname outperforms SEC by 2.07\% in R@1. These results further verify the applicability of \shortname on some regularization techniques in DML.
}

\section{Conclusion}
\lec{In this paper, we propose a \name (\shortname) scheme to alleviate the ``missing embedding'' issue incurred during DML sampling. To this end, we propose to produce embeddings with no data points by exploiting the embeddings' nearby embedding space. Specifically, we propose a \shortbfs module that identifies an embedding's discriminative features and performs random scaling on them to exploit the embedding space around it. Moreover, we propose a \shorttmb module to exploit embedding space among multiple embedding by adding the intra-class semantic differences to embeddings of the same class. By combining the embeddings with and without data points, \shortname provides more embeddings for sampling to improve the sampling quality and achieve effective DML. Extensive experiments with various loss functions and sampling methods on three public available benchmarks show that \shortname is effective. In the future, we plan to apply \shortname to other areas such as self-supervised learning that require sampling.
}

\noindent{\textbf{Acknowledgements.} This work was partially supported by Peng Cheng Laboratory Research Project No. PCL2021A07, National Natural Science Foundation of China (NSFC) 62072190, Program for Guangdong Introducing Innovative and Enterpreneurial Teams 2017ZT07X183.}


{\small
\bibliographystyle{splncs04}
\bibliography{egbib.bib}
}

\clearpage


\section*{Supplementary material}

# DAS: Densely-Anchored Sampling for Deep Metric Learning (Supplementary Materials)

Lizhao Liu[1,2], Shangxin Hunag[1], Zhuangwei Zhuang[1],
Ran Yang[1], Mingkui Tan[1,3†], and Yaowei Wang[2†]

[1]South China University of Technology  [2]PengCheng Laboratory
[3]Key Laboratory of Big Data and Intelligent Robot, Ministry of Education
{selizhaoliu,sevtars,z.zhuangwei,msyangran}@mail.scut.edu.cn,
mingkuitan@scut.edu.cn, wangyw@pcl.ac.cn

We organize our supplementary materials as follows. In Section A, we provide the detailed formulations of both pair-based and proxy-based DML loss functions. In Section B, we detail the formulations of DML sampling methods. In Section C, we provide more implementation details of DAS. In Section D, we analyze the overhead of DAS, In Section E, we provide experiment results of DAS on widely used proxy-based losses. In Section F, we provide results on DAS w/o image augmentation. In Section G, we study the effect of batch size on DAS. In Section H, we study the effect of embedding dimension on DAS. In Section I, we visualize and analyze the frequency recorder matrix. In Section J, we provide the evolution of training process w.r.t. more DML losses. In Section K, we investigate the effect of hyper-parameters $r_s, r_b, T$. In Section L, we provide qualitative results w.r.t. DFS and MTS. In Section M, we provide more qualitative results on different loss functions.

## A  Detailed Formulations of Loss Function in DML

### A.1  Pair-based Loss Function

**Contrastive Loss** [2]. The goal of contrastive loss is simply pulling the embeddings of the same class as close as possible and separating the embeddings of different classes at least of a given margin. Specifically, contrastive loss requires the index set of the sampled embedding pairs $\mathcal{P} = \{(i, j)\}$ and the pair-wise euclidean distance is calculated as $\mathbf{D}_{ij} = \|\mathbf{v}_i - \mathbf{v}_j\|$. Then the formulation of contrastive loss is as follows

$$\mathcal{L}_{\text{Contrastive}} = \sum_{(i,j)\in\mathcal{P}} \mathbb{I}\{y_i = y_j\}\, \mathbf{D}_{ij} + \mathbb{I}\{y_i \neq y_j\}\, [\gamma - \mathbf{D}_{ij}]_+ , \qquad (\text{I})$$

where $\mathbb{I}\{\cdot\}$ is the indicator function, $\gamma$ (set to 1.0 in this paper) is the margin.

**Triplet Loss** [8]. Triplet loss extends the contrastive loss by converting the absolute distance relationship between embeddings into a relative distance relationship (*i.e.,* ranking): the distance between embeddings of different classes

---

† Corresponding authors.

should be farther away than any embeddings of the same class. Specifically, triplet loss requires sampling a set of embedding triplets $\mathcal{T} = \{(a, p, n)\}$, where $y_a = y_p \neq y_n$ and $a, p, n$ are the index of the anchor, positive and negative, respectively. The formulation of triplet loss is as follows

$$\mathcal{L}_{\text{Triplet}} = \sum_{(a,p,n)\in\mathcal{T}} [\mathbf{D}_{ap} - \mathbf{D}_{an} + \gamma]_+, \tag{II}$$

where $\gamma$ (set to 0.2 in this paper) is the margin.

**Margin Loss** [11]. Margin loss introduces a more flexible optimization paradigm into the triplet loss. Specifically, a adjustable and learnable margin $\boldsymbol{\beta} \in \mathbb{R}^C$ is proposed to replace the fixed margin (*i.e.*, 0) between embedding of different classes, which converts the triplet ranking problem into a relative ordering of pairs. The formulation of margin loss is as follows

$$\mathcal{L}_{\text{Margin}} = \sum_{(i,j)\in\mathcal{P}} \mathbb{I}\{y_i = y_j\} \left[\gamma + \mathbf{D}_{ij} - \boldsymbol{\beta}_{y_i}\right]_+ + \mathbb{I}\{y_i \neq y_j\} \left[\gamma + \boldsymbol{\beta}_{y_i} - \mathbf{D}_{ij}\right]_+, \tag{III}$$

where $\gamma$ (set to 0.2 in this paper) is the margin in the triplet loss and $\boldsymbol{\beta}_{y_i}$ is the learnable margin for class $y_i$. Each element in $\boldsymbol{\beta}$ is initialized with 1.2 and the learning rate for $\boldsymbol{\beta}$ is set to $5\text{e}^{-4}$.

**Generalized Lifted Structure Loss** [3]. Generalized lifted structure loss extends the standard lifted structure loss [6] by considering all embeddings from the same class w.r.t. the anchor during intra-class distance minimization. Generalized lifted structure loss pulls embeddings of the same class w.r.t. the anchor close while pushing embeddings of different classes apart. To save computation cost, each embedding in a batch is used as the anchor once. To be specific, the index set of the sampled embeddings is $\mathcal{P} = \{(a, \mathcal{Q}, \mathcal{R})\}$, where $a \notin \mathcal{Q}, \mathcal{R}$, and $y_a = y_q \neq y_r, q \in \mathcal{Q}, r \in \mathcal{R}$. Then the formulation of generalized lifted structure loss is as follows

$$\mathcal{L}_{\text{GenLifted}} = \sum_{(a,\mathcal{Q},\mathcal{R})\in\mathcal{P}} \left[\log \sum_{q\in\mathcal{Q}} \exp\left(\mathbf{D}_{aq}\right) + \log \sum_{r\in\mathcal{R}} \exp\left(\gamma - \mathbf{D}_{ar}\right)\right]_+ + \nu \|\mathbf{v}_a\|^2, \tag{IV}$$

where $\gamma$ (set to 1.0 in this paper) is the margin to avoid pushing the embeddings of different classes too large and $\nu$ (set to $5\text{e}^{-3}$ in this paper) regularizes the embeddings. Note that, in this loss, embeddings for distance computation and producing embeddings with no data points are not normalized.

**N-Pair Loss** [9]. N-Pair loss extends the triplet loss by considering all embeddings of different classes during inter-class distance maximization. Specifically, the index set of the sampled embeddings is $\mathcal{P} = \{(a, p, \mathcal{R})\}$, where $y_a = y_p \neq y_r, r \in \mathcal{R}$, and the pair-wise distance is calculated as $\mathbf{D}_{ij} = \mathbf{v}_i^T \mathbf{v}_j$. Then the

formulation of N-Pair loss is as follows

$$\mathcal{L}_{\text{N-Pair}} = \sum_{(a,p,\mathcal{R})\in\mathcal{P}} \log\left(1 + \sum_{r\in\mathcal{R}} \exp\left(\mathbf{D}_{ar} - \mathbf{D}_{ap}\right)\right) + \nu \left\|\mathbf{v}_a\right\|^2, \qquad \text{(V)}$$

where $\nu$ (set to $5\mathrm{e}^{-3}$ in this paper) controls the optimization strength on the embedding regularization. Note that, in this loss, embeddings for distance computation and producing embeddings with no data points are also not normalized.

**Multi-similarity Loss** [10]. Apart from considering simple anchor-positive, anchor-negative relationships, multi-similarity loss better leverages all embeddings in a batch by additionally considering positive-positive and negative-negative relationship. Also, to save computation cost, each embedding in a batch will only be used as anchor once. For a anchor $a$, let $\mathcal{P}_a$ and $\mathcal{N}_a$ denote its corresponding positive and negative embedding index sets, respectively. Given the pair-wise distance computed by $\mathbf{D}_{ij} = \mathbf{v}_i^T \mathbf{v}_j$, the sampled embedding index set is constructed as $\mathcal{P} = \{(a, \mathcal{Q}, \mathcal{R})\}$, where $a \notin \mathcal{Q}, \mathcal{R}$, $\mathcal{Q} = \{q \ | \ y_q = y_a, \ \mathbf{D}_{aq} > \min_{i\in\mathcal{P}_a}(\mathbf{D}_{ai} - \epsilon)\}$ and $\mathcal{R} = \{r \ | \ y_r \neq y_a, \ \mathbf{D}_{ar} < \max_{j\in\mathcal{N}_a}(\mathbf{D}_{aj} + \epsilon)\}$. Then the formulation of multi-similarity loss is as follows

$$\begin{aligned}\mathcal{L}_{\text{MS}} = \sum_{(a,\mathcal{Q},\mathcal{R})\in\mathcal{P}} &\frac{1}{\alpha} \log\left[1 + \sum_{q\in\mathcal{Q}} \exp\left(-\alpha\left(\mathbf{D}_{aq} - \lambda\right)\right)\right] \\ &+ \frac{1}{\beta} \log\left[1 + \sum_{r\in\mathcal{R}} \exp\left(\beta\left(\mathbf{D}_{ar} - \lambda\right)\right)\right],\end{aligned} \qquad \text{(VI)}$$

where $\alpha, \beta, \lambda, \epsilon$ are hyper-parameters to be set. In this paper, we set $\alpha = 2, \beta = 40, \lambda = 5\mathrm{e}^{-1}, \epsilon = 1\mathrm{e}^{-1}$.

### A.2  Proxy-based Loss Function

**Softmax Loss** [12]. Different from the pair-based loss function, softmax loss1 introduces a proxy *i.e.,* classifier for each class and optimizes the embedding by pulling it close to its proxy. The formulation of softmax loss is as follows:

$$\mathcal{L}_{\text{Softmax}} = -\sum_i \log \frac{\exp\left(\mathbf{W}_{y_i}^T \mathbf{v}_i \ / \ T\right)}{\sum_{c\in C} \exp\left(\mathbf{W}_c^T \mathbf{v}_i \ / \ T\right)}, \qquad \text{(VII)}$$

where $\mathbf{W} \in \mathbb{R}^{C\times d}$ is the classifier weight for all training classes. Since the embedding $\mathbf{v}_i$ is normalized, a temperature $T$ (set to $5\mathrm{e}^{-2}$ in this paper) is used to boost the gradient. Moreover, the learning rate of $\mathbf{W}$ is set to $1\mathrm{e}^{-5}$ for CARS and CUB, $2\mathrm{e}^{-3}$ for SOP.

**ArcFace Loss** [1]. ArcFace loss improves the vanilla softmax by adding an angular margin into embedding and its corresponding proxy to achieve more

compact intra-class representation. The formulation of ArcFace loss is as follows

$$\mathcal{L}_{\text{ArcFace}} = -\sum_i \log \frac{\exp\left(s \cdot \cos\left(\mathbf{W}_{y_i}^T \mathbf{v}_i + \gamma\right)\right)}{\exp\left(s \cdot \cos\left(\mathbf{W}_{y_i}^T \mathbf{v}_i + \gamma\right)\right) + \sum_{c \neq y_i} \exp\left(s \cdot \cos\left(\mathbf{W}_c^T \mathbf{v}_i\right)\right)}, \tag{VIII}$$

where $\mathbf{W} \in \mathbb{R}^{C \times d}$ is the classifier weight for all training classes and $\gamma, s$ are hyperparameters to be set. In this paper, we set $\gamma = 5e^{-1}$ and $s = 16$. Moreover, the learning rate of $\mathbf{W}$ is set to $5e-4$ for all datasets.

## B   Detailed Formulations of Sampling Method for DML

**Random Sampling** [4]. Random sampling simply selects the index of positive pair or negative pair in a most trivial way *i.e.*, randomly selecting. To be specific, given an embedding $\mathbf{v}_i$, its index of positive is randomly draw from $\{j \mid y_i = y_j, i \neq j\}$ and its index of negative is randomly draw from $\{k \mid y_i \neq y_k, i \neq k\}$.

**Semi-hard Sampling** [8]. Semi-hard sampling is proposed to effectively sample embedding triplets that grows cubically to batch size. In the training process, most of the triplets satisfy the objective function and they provide limited (or no) training signal to train the model, thereby impeding the model learning [8]. Thus, given an anchor $\mathbf{v}_a$ and its positive $\mathbf{v}_p$ (randomly sampled), semi-hard sampling carefully choose negative embedding's index as follows

$$n \sim \{i \mid y_i \neq y_a, \|\mathbf{v}_a - \mathbf{v}_p\|^2 < \|\mathbf{v}_a - \mathbf{v}_i\|^2\}. \tag{IX}$$

**Soft-hard Sampling** [7] To avoid selecting "hard" embeddings that impedes model training, semi-hard sampling chooses embeddings that are relatively close to the anchor. Soft-hard triplet sampling shows that a probabilistic (soft) selection of potentially hard embeddings is actually beneficial. Given an anchor embedding $\mathbf{v}_a$, soft-hard sampling attain the indexes of positive and negative embedding as follows

$$p \sim \{i \mid y_i = y_a, \|\mathbf{v}_a - \mathbf{v}_i\|^2 > \arg\min_{q \in \mathcal{Q}_a} \|\mathbf{v}_a - \mathbf{v}_q\|^2\}, \tag{X}$$

$$n \sim \{j \mid y_j \neq y_a, \|\mathbf{v}_a - \mathbf{v}_j\|^2 < \arg\max_{r \in \mathcal{R}_a} \|\mathbf{v}_a - \mathbf{v}_r\|^2\}, \tag{XI}$$

where $\mathcal{R}_a = \{r \mid y_r \neq y_a\}$, $\mathcal{Q}_a = \{q \mid y_q \neq y_a\}$ are the positive and negative index sets w.r.t. the anchor $a$, respectively. In this way, soft-hard sampling explores more triplets than semi-hard sampling to improve the model training.

**Distance-weighted Sampling** [11] . Different from other sampling strategy that considers a certain distance range of embeddings, distance-weighted sampling considers a wide range of embeddings in a probabilistic way. Since the

embedding space is typically a $d$-dimensional hypersphere $\mathbb{S}^{d-1}$, the analytical distribution of pairwise distance on a hypersphere obeys

$$q\left(\mathbf{D}_{ij}\right) \propto \mathbf{D}_{ij}^{d-2}[1 - \frac{1}{4}\mathbf{D}_{ij}]^{\frac{d-3}{2}}, \qquad \text{(XII)}$$

and $\mathbf{D}_{ij} = \|\mathbf{v}_i - \mathbf{v}_j\|$ for any embedding pairs $\mathbf{v}_i, \mathbf{v}_j \in \mathbb{S}^{d-1}$. To obtain a wide range of negative embeddings that are able to improve the embedding diversity as well as model training, distance-weighted sampling acquires the index of negative embedding based on the inversed distance distribution

$$P(n \mid a) \propto \min\left(\lambda, q^{-1}(\mathbf{D}_{an})\right). \qquad \text{(XIII)}$$

In this paper, we set $\lambda = 5\mathrm{e}^{-1}$ and the largest distance to 1.4.

## C   More Implementation Details

In this section, we provide more implementation details. As for image augmentation process, random crop (image size 224×224) with random horizontal flip ($p = 0.5$) is applied during training and single center crop (image size 256×256) is used for testing. In terms of training strategy, the number of training epochs is 300. We use Adam [5] as the optimizer. The initial learning rate is $1\mathrm{e}^{-5}$, which is reduced by a factor of 0.3 in $200^{\text{th}}$ and $250^{\text{th}}$ epoch, respectively. The weight decay is $4\mathrm{e}^{-4}$. For batch preparation, SPC-2 construction [7] is used (2 samples per category). The batch size is set to 112.

## D   Efficiency and Overhead Analysis

DAS takes extra cost only in the training stage. Specifically, w/ and w/o DAS, the training time cost for [11] are 1.15s *vs.* 0.70s per batch, which includes the cost of DAS and using more embeddings for sampling and loss computation. Moreover, DAS only consumes 13% of the total time, which is efficient compared to the whole training procedure.

## E   Effectiveness of DAS on Proxy-based Loss

Although DAS is developed for pair-based loss, we perform experiments to evaluate the generalization ability of DAS on classic and widely used proxy-based losses *i.e.,* Softmax and ArcFace. The results are presented in Table I. The improvements are still observed when equipped with DAS for Softmax and ArcFace across different datasets.

## F   DAS w/o Image Augmentation

We further perform experiments without image augmentation using triplet loss and distance weighted sampling on CARS. The results are shown in Table II. DAS boosts all metrics considerably, showing that DAS is complementary to image augmentation technique.

Table I: Comparisons with proxy-based approaches on various datasets

| Method | CUB | | | CARS | | | SOP | | |
|---|---|---|---|---|---|---|---|---|---|
| | R@1 | F1 | NMI | R@1 | F1 | NMI | R@1 | F1 | NMI |
| Softmax [12] | 61.58 | 36.12 | 66.73 | 79.07 | 37.11 | 67.01 | 77.92 | 37.20 | 90.05 |
| Softmax + DAS | **62.02** | **36.24** | **67.42** | **81.23** | **39.95** | **68.91** | **79.36** | **38.72** | **90.40** |
| ArcFace [1] | 61.56 | 35.73 | 66.83 | 79.50 | 37.75 | 67.82 | 78.08 | 37.79 | 90.18 |
| ArcFace + DAS | **62.80** | **37.63** | **67.80** | **82.22** | **40.82** | **69.82** | **78.12** | **38.08** | **90.26** |

Table II: Comparisons without image augmentation on CARS

| DAS | R@1 | F1 | NMI |
|---|---|---|---|
| | 61.47 | 22.06 | 53.88 |
| ✓ | 65.13 (**+3.66**) | 23.77 (**+1.71**) | 55.89 (**+2.01**) |

## G  Effect of Batch Size

In this section, we investigate the effect of batch size on the proposed DAS. The results are presented in Fig. I. The loss function and sampling method are margin loss [11] and distance-weighted sampling [11], respectively. From Fig. I, we have the following observations: **First**, under various batch size and image retrieval evaluation metrics, when equipped with DAS, the model is able to consistently obtain better results than the one trained without DAS. **Second**, we observe that the model trained with DAS and batch size = 32 outperforms the one trained without DAS and batch size = 224 in terms of R@1. It shows that producing effective embeddings without datapoints by DAS is as equally important as providing more data points in a batch to achieve improved performance. These results well prove the rationality of our motivation and the efficacy of DAS.

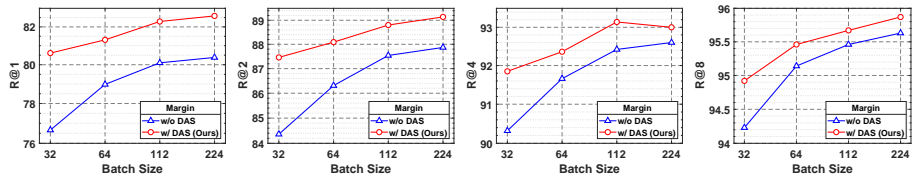

Fig. I: The test set R@$\{1, 2, 4, 8\}$ on CARS with different batch size

## H  Effect of Embedding Dimension

In this section, we evaluate the proposed method on different embedding dimensions. The results are shown in Fig. II. We use the margin loss [11] as the

loss function while leveraging the distance-weighted sampling as the sampling method [11]. From Fig. II, we obtain the following results: **First**, under different embedding dimension, DAS consistently reaches the best performance for all image retrieval metrics. **Second**, the model that trained with DAS and embedding dimension = 64 obtains a comparable result like the one trained without DAS and embedding dimension = 128 regarding R@1. It shows that the produced embeddings by DAS are able to force the model to better leverage the model capacity. And covering the barren area in embedding space is important to get improved performance when model capacity is low. These results demonstrate the effectiveness of DAS.

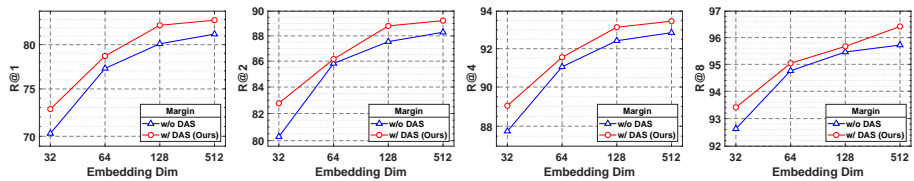

(a) Experiments using different embedding dim.

Fig. II: Test set R@$\{1, 2, 4, 8\}$ on CARS with different embedding dimension $d$

## I  Visualization Results on Frequency Recorder Matrix

In this section, we visualize the Frequency Recorder Matrix (FRM) **P** introduced in the DFS module. The FRM serves as a stable and effective identifier for semantic scaling by considering the top activated features for one class as the effective semantics instead of noises. The loss function and sampling method we used here are triplet loss [8] and distance-weighted sampling [11], respectively. We perform experiments on all three datasets (*i.e.,* CARS, CUB and SOP). The results are depicted in Fig. III, from which, we have the following observations: **First**, for different training stages (*i.e.,* epoch = 1, 150, 300), the number of top activated features for embeddings of the same classes are limited (*i.e.,*around $4 \sim 8$) across all datasets. **Second**, as the training process proceeds, more features are likely to be the top activated features. **Third**, for the large scale dataset SOP, more features are likely to be the top activated ones due to the rich semantics covered by adequate data points. In this sense, with the proposed FRM, we are able to figure out channels with more discriminative power to achieve effective semantic scaling. These results demonstrate the rationality of the proposed FRM.

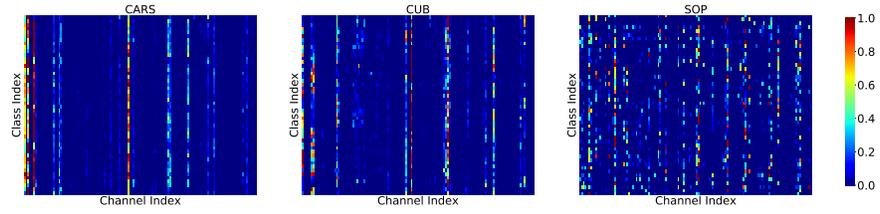

(a) Visualization of the frequency recorder matrix at epoch = 1

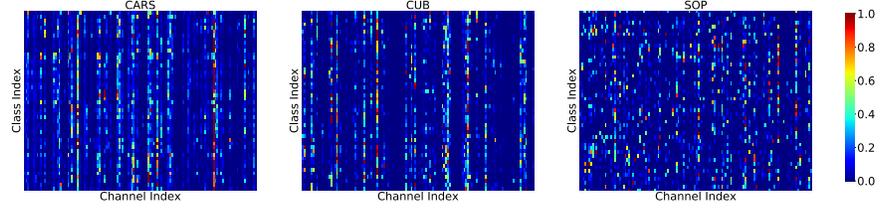

(b) Visualization of the frequency recorder matrix at epoch = 150

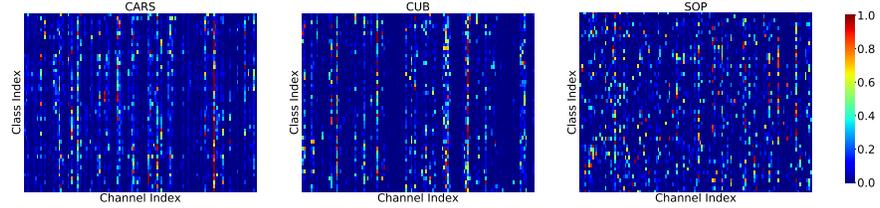

(c) Visualization of the frequency recorder matrix at epoch = 300

Fig. III: From left to right, the visualized FRM on CARS, CUB and SOP, respectively. Each element in $\mathbf{P}$ is normalized (*i.e.,* divided by the maximum value in its row). Only the first 48 classes are presented due to page limit

## J   Evolution of Training Process w.r.t. Different Losses

In this section, we provide the evolution of training loss and test set R@1 w.r.t. different losses in the training process. The results are depicted in Fig. IV. We have the following observations: **First**, when training with DAS, the training losses are generally higher and decrease smoother than the baseline, which demonstrates that producing more embeddings by DAS is able to consistently provide training signal to train the model. **Second**, when equipped with DAS, the test set R@1s are higher than the baseline. **Third**, for some loss functions that face severe overfitting problems such as contrastive loss and generalized lifted structure loss, DAS is able to ease the overfitting problem. These results verify the effectiveness of the proposed method across different loss functions and sampling methods.

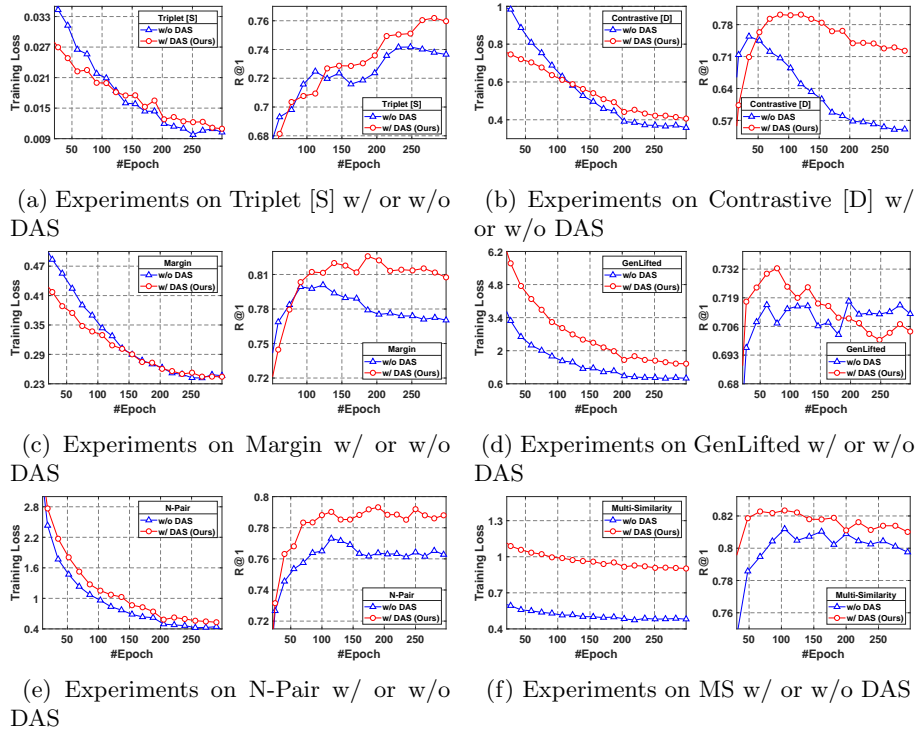

(a) Experiments on Triplet [S] w/ or w/o DAS

(b) Experiments on Contrastive [D] w/ or w/o DAS

(c) Experiments on Margin w/ or w/o DAS

(d) Experiments on GenLifted w/ or w/o DAS

(e) Experiments on N-Pair w/ or w/o DAS

(f) Experiments on MS w/ or w/o DAS

Fig. IV: The training loss and test set R@1 on CARS with different losses

## K    Ablation Studies on Hyper-parameters

In this section, we investigate the effect of the hyper-parameters $r_s, r_b, T$ in DAS. The loss function and sampling method are margin loss [11] and distance-weighted sampling [11], respectively. The default hyper-parameters' settings are $(r_s, r_b, T) = (1e^{-2}, 1e^{-2}, 3)$.

**Random scale in DFS** ($r_s$). The results on different random scale in DFS are shown in Table III (a). Our method is insensitive to a wide range of random scale, showing that scaling the discriminative features is able to provide effective semantics of different strength.

**Semantic shifting scale in MTS** ($r_b$). $r_b$ is to provide the flexibility of controlling the strength of adding intra-class semantic differences. The results on different semantic shifting scales in MTS are shown in Table III (b). Our method obtains similar results under different $r_b$ and reaches the best results when $r_b = 1$, which suggests that larger $r_b$ is able to cover more barren area in the embedding space to improve the model training.

**Number of the produced embeddings** $T$. The results on different numbers of the produced embeddings are shown in Table III (c). As $T$ increases from 1 to 5, the proposed DAS achieves better results and reaches the best result at $T = 5$. When $T = 10$, the performance is worse than $T = 5$, which indicates that too many produced embeddings with no data points will dominate the optimization direction and impair the learning of embeddings with data points.

Table III: Experiments on different hyper-parameters on CARS

(a) Effect of the random scale $r_s$ in DFS

| $r_s$ | $1e^{-2}$ | $1e^{-1}$ | $2e^{-1}$ | $5e^{-1}$ |
|---|---|---|---|---|
| R@1 | 82.29 | 82.25 | 82.30 | **82.43** |

(b) Effect of the scale $r_b$ in MTS

| $r_b$ | $1e^{-3}$ | $1e^{-2}$ | $1e^{-1}$ | 1 |
|---|---|---|---|---|
| R@1 | 82.19 | 82.29 | 82.07 | **82.55** |

(c) Effect of the number of produced embedding ($T$) in DAS

| $T$ | 1 | 3 | 5 | 10 |
|---|---|---|---|---|
| R@1 | 80.78 | 82.29 | **83.40** | 81.75 |

## L    Qualitative Results of DFS and MTS

In this section, we investigate the effectiveness of the proposed DFS and MTS. Specifically, we apply DFS and MTS in the test phase and compare results from the model trained with or without them. Since the training and test classes are different, the DFS and MTS modules used for training are unavailable here.

Thus, the semantic scaling is implemented as randomly scaling the top $K$ features in an embedding; Whilst we perform semantic shifting by adding the transformation (obtained from another two embeddings of the same class) to the embedding. The loss function and sampling method we used here are contrastive loss [2] and distance-weighted sampling [11], respectively. The results for semantic scaling and shifting are shown in Fig. V and Fig. VI, respectively. We have the following observations: **1)** When we apply different semantic scaling to the query, the model trained with DAS consistently retrieves correct results, which is not the case for the baseline. **2)** The model trained with DAS is able to retrieve expected results even with the semantic shifted embedding while the baseline fails to do so. These results show that DAS is able to produce embeddings with effective semantics to train the model, which is insensitive to the semantic differences and consistently achieves good generalization ability after training.

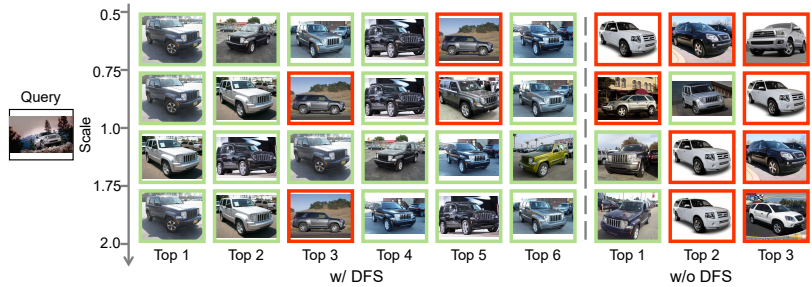

Fig. V: Top 6 retrieved results with different scales on CARS. The expected and unexpected results are framed by green and red rectangles, respectively

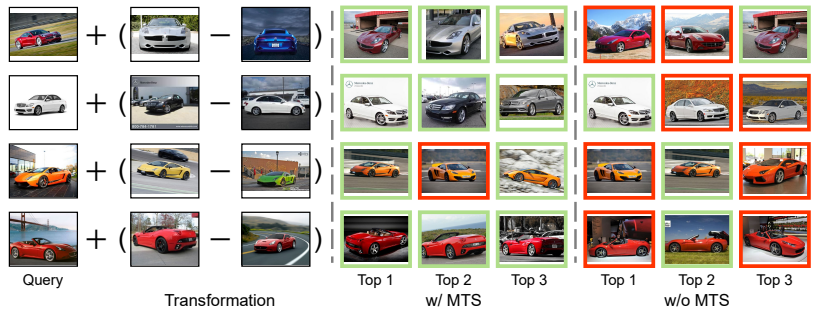

Fig. VI: Top 3 retrieved results with MTS on CARS. The expected and unexpected results are framed by green and red rectangles, respectively

## M    More Qualitative Results

In this section, we provide qualitative results on different losses w/ or w/o DAS. The results for CARS and SOP are in Fig. VII and Fig. VIII, respectively. From those results, we can see that the proposed DAS can enforce the model to focus on real semantics despite the background noises and other semantics' interference such as car's colors, drastic viewpoint changes *etc*. These results show the generalization ability and robustness of the proposed DAS.

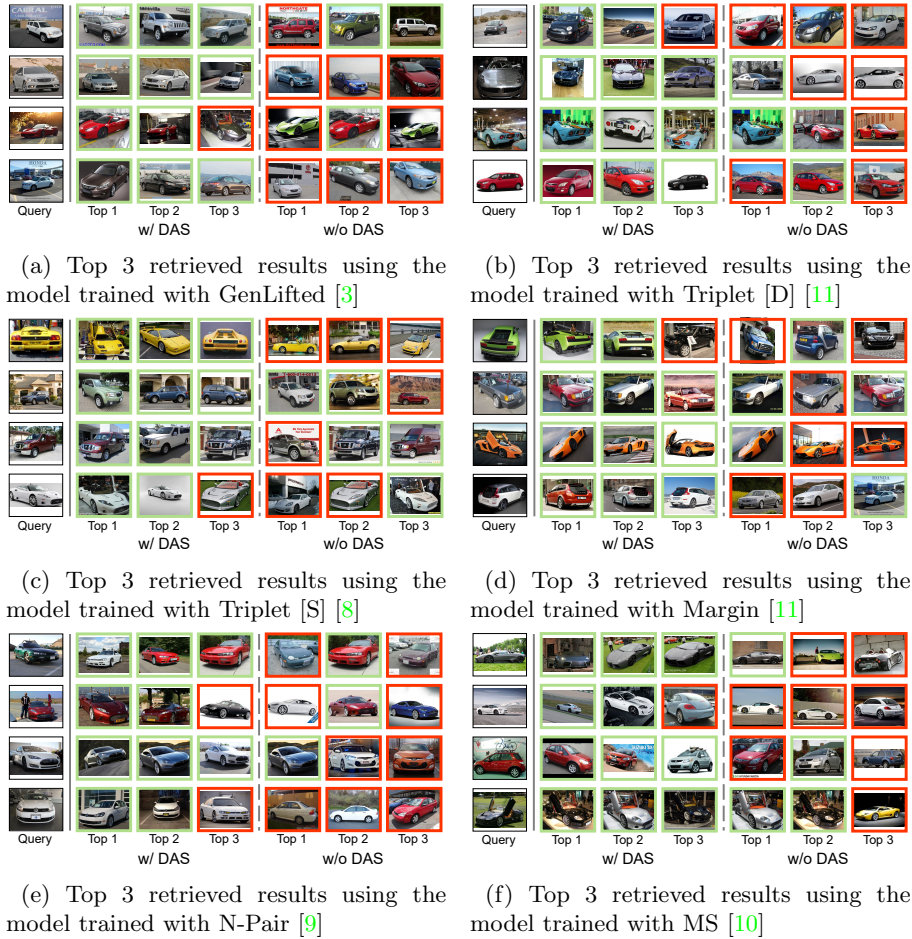

(a) Top 3 retrieved results using the model trained with GenLifted [3]

(b) Top 3 retrieved results using the model trained with Triplet [D] [11]

(c) Top 3 retrieved results using the model trained with Triplet [S] [8]

(d) Top 3 retrieved results using the model trained with Margin [11]

(e) Top 3 retrieved results using the model trained with N-Pair [9]

(f) Top 3 retrieved results using the model trained with MS [10]

Fig. VII: Top 3 retrieved results using the model trained by different loss functions that are equipped w/ or w/o on CARS. The expected and unexpected results are framed by green and red rectangles, respectively

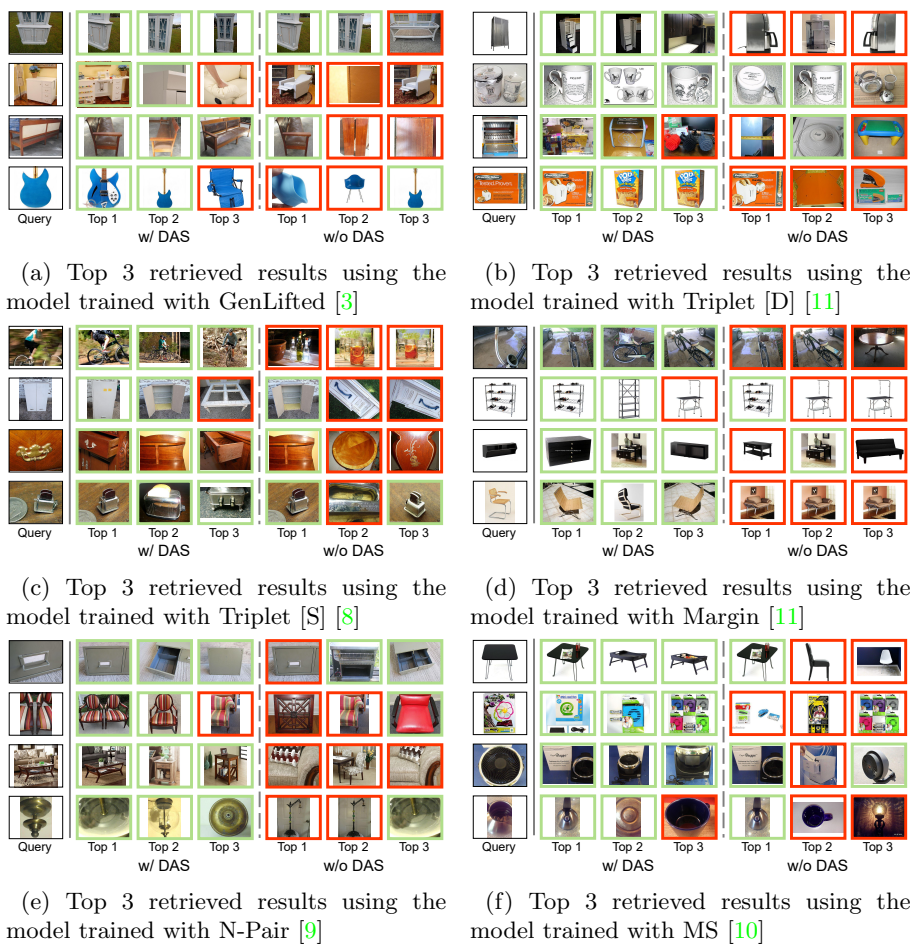

(a) Top 3 retrieved results using the model trained with GenLifted [3]

(b) Top 3 retrieved results using the model trained with Triplet [D] [11]

(c) Top 3 retrieved results using the model trained with Triplet [S] [8]

(d) Top 3 retrieved results using the model trained with Margin [11]

(e) Top 3 retrieved results using the model trained with N-Pair [9]

(f) Top 3 retrieved results using the model trained with MS [10]

Fig. VIII: Top 3 retrieved results using the model trained by different loss functions that are equipped w/ or w/o on SOP. The expected and unexpected results are framed by green and red rectangles, respectively



\end{document}